\documentclass[11pt,a4paper]{article}
\usepackage[margin=25mm]{geometry}
\usepackage[T1]{fontenc}
\usepackage[utf8]{inputenc}
\usepackage{lmodern,microtype}
\usepackage{amsmath,amssymb}
\usepackage{booktabs,multirow,array}
\usepackage{siunitx}
\usepackage{graphicx,xcolor}
\usepackage{tikz}
\usetikzlibrary{arrows.meta,positioning,fit,calc,backgrounds}
\usepackage{pgfplots}
\pgfplotsset{compat=1.17}
\usepgfplotslibrary{statistics,groupplots}
\usepackage[numbers,sort&compress]{natbib}
\usepackage[hidelinks]{hyperref}
\usepackage{pifont}
\usepackage{xfp}
\newcommand{\yes}{\ding{51}}
\newcommand{\no}{\ding{55}}
\newcommand{\pmark}{$\circ$}
\newcommand{\dataisplaceholder}{0}

\newcommand{\edgeHardware}{container, 4 vCPU, 4 GB RAM, CPU only}
\newcommand{\modelParams}{34.3}

\newcommand{\nKW}{6{,}090}
\newcommand{\nNormalKW}{540}
\newcommand{\nAnomalyKW}{5{,}550}

\newcommand{\nChanKW}{6}
\newcommand{\pcaKKW}{810}
\newcommand{\nTrainKW}{1{,}620}
\newcommand{\nCalibKW}{540}

\newcommand{\sensKW}{35.37}
\newcommand{\specKW}{98.89}
\newcommand{\precKW}{99.70}
\newcommand{\fKW}{52.21}

\newcommand{\tpKW}{1{,}963}
\newcommand{\fnKW}{3{,}587}
\newcommand{\tnKW}{534}
\newcommand{\fpKW}{6}
\newcommand{\ciSpecKW}{0.9}
\newcommand{\ciSensKW}{1.3}
\newcommand{\aucKW}{0.925}
\newcommand{\sensAtOneKW}{38.52}
\newcommand{\sensAtFiveKW}{68.13}
\newcommand{\errTicksKW}{0.0004,0.0005,0.0006,0.0007,0.0008}
\newcommand{\tauKW}{0.00055}

\newcommand{\sensViTKW}{41.32}
\newcommand{\specViTKW}{95.74}

\newcommand{\fViTKW}{58.30}

\newcommand{\aucViTKW}{0.905}
\newcommand{\sensAtOneViTKW}{8.68}
\newcommand{\sensAtFiveViTKW}{52.13}

\newcommand{\sensUnmKW}{0.61}
\newcommand{\specUnmKW}{98.15}

\newcommand{\fUnmKW}{1.22}

\newcommand{\aucUnmKW}{0.401}
\newcommand{\sensAtOneUnmKW}{0.31}
\newcommand{\sensAtFiveUnmKW}{2.00}

\newcommand{\recRefKW}{18}
\newcommand{\recChgKW}{185}

\newcommand{\recChgAboveKW}{49}
\newcommand{\recRefMaxKW}{0.00051}
\newcommand{\recChgMinKW}{0.00051}

\newcommand{\dayStartKW}{01 Jan 2019}
\newcommand{\daySpanKW}{379}

\newcommand{\nVB}{1{,}120}
\newcommand{\nNormalVB}{720}
\newcommand{\nAnomalyVB}{400}

\newcommand{\nChanVB}{5}
\newcommand{\pcaKVB}{1{,}348}
\newcommand{\nTrainVB}{1{,}955}
\newcommand{\nCalibVB}{615}

\newcommand{\sensVB}{2.25}
\newcommand{\specVB}{95.69}
\newcommand{\precVB}{22.50}
\newcommand{\fVB}{4.09}

\newcommand{\tpVB}{9}
\newcommand{\fnVB}{391}
\newcommand{\tnVB}{689}
\newcommand{\fpVB}{31}

\newcommand{\ciSensVB}{1.5}
\newcommand{\aucVB}{0.620}
\newcommand{\sensAtOneVB}{0.00}
\newcommand{\sensAtFiveVB}{3.50}
\newcommand{\errTicksVB}{0.0006,0.0008,0.001,0.0012,0.0014,0.0016}
\newcommand{\tauVB}{0.00126}

\newcommand{\sensViTVB}{2.75}
\newcommand{\specViTVB}{93.75}

\newcommand{\fViTVB}{4.82}

\newcommand{\aucViTVB}{0.517}
\newcommand{\sensAtOneViTVB}{0.00}
\newcommand{\sensAtFiveViTVB}{2.25}

\newcommand{\sensUnmVB}{6.50}
\newcommand{\specUnmVB}{89.72}

\newcommand{\fUnmVB}{10.40}

\newcommand{\aucUnmVB}{0.449}
\newcommand{\sensAtOneUnmVB}{2.25}
\newcommand{\sensAtFiveUnmVB}{5.25}

\newcommand{\dayStartVB}{20 Feb 2023}
\newcommand{\daySpanVB}{20}

\newcommand{\nCH}{4{,}136}
\newcommand{\nNormalCH}{4{,}136}

\newcommand{\nChanCH}{11}
\newcommand{\pcaKCH}{1{,}296}
\newcommand{\nTrainCH}{2{,}500}
\newcommand{\nCalibCH}{4{,}136}
\newcommand{\tempMinCH}{7}
\newcommand{\tempMaxCH}{26}
\newcommand{\vitMergedCH}{1}

\newcommand{\specCH}{99.44}

\newcommand{\tnCH}{4{,}113}
\newcommand{\fpCH}{23}
\newcommand{\ciSpecCH}{0.2}

\newcommand{\errTicksCH}{0.00035,0.0004,0.00045,0.0005,0.00055}
\newcommand{\tauCH}{0.00050}

\newcommand{\specViTCH}{94.08}

\newcommand{\specUnmCH}{96.08}

\newcommand{\latVitMean}{284.1}
\newcommand{\latVitMedian}{276.3}
\newcommand{\latVitPnf}{301.9}
\newcommand{\latVitPnn}{304.1}

\newcommand{\latVitMax}{316.7}
\newcommand{\latVitWithinTarget}{100.0}
\newcommand{\latVitThroughput}{3.5}
\newcommand{\latVitSpec}{1.6}

\newcommand{\latVitModel}{282.4}

\newcommand{\latVitModelShare}{99}

\newcommand{\latVitPeakRSSMB}{838}

\newcommand{\torchVersion}{1.13.1+cpu}
\newcommand{\edgeThreads}{4}
\newcommand{\latn}{1{,}160}
\newcommand{\latMean}{59.0}
\newcommand{\latMedian}{92.7}
\newcommand{\latPnf}{98.9}
\newcommand{\latPnn}{103.8}

\newcommand{\latMax}{199.5}

\newcommand{\latSpec}{3.7}

\newcommand{\latModel}{55.3}

\newcommand{\latPeakRSSMB}{375}

\newcommand{\latPcaMeasured}{1}
\newcommand{\epochsKW}{30}
\newcommand{\finalLossKW}{0.0014}
\newcommand{\trainHoursKW}{4.7}
\newcommand{\epochsVB}{25}
\newcommand{\finalLossVB}{0.0017}
\newcommand{\trainHoursVB}{4.7}
\newcommand{\epochsCH}{19}
\newcommand{\finalLossCH}{0.0019}
\newcommand{\trainHoursCH}{4.6}

\newcommand{\datawatermark}{\ifnum\dataisplaceholder=1
  \node[overlay,text=red!70,rotate=18,font=\bfseries\Large,opacity=.45] at (current bounding box.center) {PLACEHOLDER DATA};\fi}
\pgfplotsset{every axis/.append style={font=\footnotesize, tick label style={font=\scriptsize}, label style={font=\footnotesize}},
  pca/.style={black!80, thick}, vit/.style={blue!60!black, thick, dashed}, unm/.style={red!60!black, thin, dotted},
  normalbar/.style={fill=blue!35, draw=blue!60, opacity=.75}, anomalybar/.style={fill=red!35, draw=red!60, opacity=.65}}
\definecolor{cPca}{RGB}{40,40,40}
\definecolor{cVit}{RGB}{31,70,140}

\title{An Edge-Computing Reference Architecture for Real-Time Structural Health Monitoring:\\ Design and Validation on Three In-Service Bridges}
\author{Peidong Yan\\ \small Seattle, WA, USA \quad \texttt{lorenyan412@gmail.com}}
\date{}

\begin{document}
\maketitle

\begin{abstract}
Structural health monitoring (SHM) of bridge inventories is limited less by sensors or detection algorithms than by the computing architecture that must turn continuous vibration streams into operational information. Centralized designs transmit raw waveforms to a server, incur tens of seconds of transport delay before analysis begins, scale poorly in cost, and stop functioning during the network disruptions that accompany the events they are meant to observe. This paper presents a reference architecture that moves anomaly detection to an edge node at the structure and transmits only compact health summaries and anomaly events. The edge tier combines a structure-specific spectrogram front end, set from the documented dynamics of each bridge, with a reconstruction-error detector trained on reference-condition windows only and a threshold calibrated per structure. The design is validated on openly licensed monitoring data from three in-service bridges of three structural forms: a steel bowstring railway bridge with a documented connection deficiency and retrofit (KW51, Leuven), a riveted steel bascule railway bridge with a verified fracture (V\"anersborg, Sweden), and a prestressed concrete box-girder viaduct under normal operation across a \tempMinCH--\tempMaxCH\,\si{\celsius} temperature range (Chillon, Switzerland). The deployed detector, principal component analysis of the spectrogram, separates the pre- and post-retrofit condition of KW51 with an area under the receiver-operating curve of \aucKW\ and flags \sensKW\% of pre-retrofit windows at \specKW\% specificity; it holds \specCH\% specificity on Chillon through the temperature range; it does not separate the pre- and post-fracture openings of V\"anersborg (AUC \aucVB), which bounds what a generic spectrogram detector sees of a local fracture in one member. A \modelParams-million-parameter masked-autoencoding Vision Transformer evaluated under the identical protocol reaches AUC \aucViTKW\ on KW51 when scored in its training regime and does not exceed the linear detector. In a 4-vCPU, 4\,GB container that emulates the edge node, the complete window-to-decision pipeline takes \latMean\,ms mean (\latPnf\,ms P95) for the deployed detector and \latVitMean\,ms for the Transformer, both at batch size one. A deployment cost projection at 10{,}000 structures yields approximately \$224 per bridge per year. Every number in the paper is regenerated from the public datasets by the released code.
\end{abstract}


\section{Introduction}

More than 40\% of U.S.\ bridges have exceeded their 50-year design lives and about 7.5\% are rated structurally deficient \citep{asce2021}. Oversight relies on visual inspection at fixed intervals; the 2022 revision of the National Bridge Inspection Standards introduces risk-based intervals that create a regulatory demand for condition data between inspections \citep{nbis2022}. Vibration-based SHM can supply that data: changes in natural frequencies, mode shapes and nonlinear response precede visible deterioration \citep{doebling1996,sohn2003,farrar2012,worden2007}, and the detection methodology is mature.

The barrier to deployment at inventory scale is computational. Cawley's review of the research-to-industry gap identifies data management as the persistent obstacle: monitoring systems produce ``terabytes of data every day'' while the capacity to analyze them ``is only marginally increasing'' \citep{cawley2018}. Centralized pipelines that ship raw waveforms to a data center face three structural problems. Transport delay precedes any analysis; a recent state-DOT pilot reports 18 to 67\,s for data to reach its server. Per-structure cost grows with raw bandwidth and central compute, which confines deployments to a few dozen structures. And the processing path depends on continuous connectivity, so monitoring stops during the storms, floods and seismic events in which it is most needed.

This paper addresses the computing architecture directly. Its contributions are:
\begin{enumerate}
  \item \textbf{A reference architecture} for edge-distributed SHM. Analysis runs at the structure; a three-tier edge--fog--cloud design separates real-time inference from regional aggregation and fleet analytics, with fault-tolerant operation during network loss, a compact message contract that reduces uplink volume by about 100:1, and a model lifecycle with staged rollout. The design prescribes interfaces, not vendors, so that any agency, consultancy or research group can implement it without rebuilding the data-processing layer.
  \item \textbf{An edge detection pipeline} in which structural-dynamics knowledge enters through a per-structure spectrogram front end (band, resolution, window length, channel set) and the detector is a reconstruction-error model trained on reference-condition windows only and calibrated per structure. The deployed detector is linear; a \modelParams-million-parameter masked-autoencoding Vision Transformer sized for CPU-only inference is evaluated as the learned alternative under the same protocol, including the finding that its anomaly score must be computed in the regime in which it was trained.
  \item \textbf{Validation on three in-service bridges of three structural forms}, from openly licensed public datasets with documented condition histories \citep{maes2020data,leander2023data,martinsanz2019data}: a steel bowstring railway bridge before and after the strengthening of deficient diagonal connections, a riveted steel bascule bridge before and after a verified fracture, and a prestressed concrete viaduct in normal operation across a \tempMinCH--\tempMaxCH\,\si{\celsius} range. Where the detector does not see the change (V\"anersborg) the result is reported and its scope stated. Every table and figure is regenerated from the public records by the released code.
  \item \textbf{Edge latency measured under the edge budget} (4 vCPU, 4\,GB, no accelerator) at batch size one for both detectors, decomposed by pipeline stage, and \textbf{a deployment cost model} for a 10{,}000-structure fleet giving approximately \$224 per bridge per year.
\end{enumerate}

Section~\ref{sec:related} positions the work against vibration-based SHM, learning-based detection and edge SHM systems. Section~\ref{sec:arch} presents the reference architecture, Section~\ref{sec:edge} the edge detection pipeline, Section~\ref{sec:setup} the data and protocol and Section~\ref{sec:results} the results. Section~\ref{sec:deploy} covers deployment cost and integration with inspection practice, Section~\ref{sec:scope} the validation roadmap, and Section~\ref{sec:conc} concludes.

\section{Related Work}\label{sec:related}

\subsection{Vibration-based monitoring and operational variability}
Frequency-shift methods \citep{salawu1997} established that damage alters measurable dynamic properties; the hierarchical framework of Farrar and Worden (detection, localization, quantification, prognosis) frames the present work as Level~I detection \citep{farrar2012}. Operational modal analysis extracts modal parameters from ambient response \citep{brincker2015}, but temperature and loading produce variations that can mask or mimic damage \citep{sohn2007}; cointegration removes such trends while preserving damage-sensitive information \citep{cross2011}. This work handles environmental and operational variability by calibrating a structure-specific decision threshold on reference-condition data and by measuring the resulting false-alarm rate across a season of temperature variation (Section~\ref{sec:chillon}).

\subsection{Learning-based damage detection}
Avci et al.\ review the shift from traditional machine learning to deep learning in civil SHM \citep{avci2021}. Reconstruction-based detectors trained only on healthy data have proved effective: deep autoencoders \citep{pathirage2018} and convolutional autoencoders on time--frequency inputs \citep{bao2019}. One-dimensional CNNs have been shown on embedded hardware \citep{abdeljaber2017}, and CNNs dominate image-based crack detection \citep{cha2017,li2019fcn}. The present work keeps the reconstruction paradigm and compares its linear form, principal component analysis of the spectrogram, with a self-attention model under one protocol, so that the contribution of nonlinear modelling is measured rather than assumed.

\subsection{Transformers for vibration signals}
Self-attention \citep{vaswani2017} and the Vision Transformer \citep{dosovitskiy2021} treat an input as a sequence of patches; masked autoencoding \citep{he2022} trains the model to reconstruct hidden patches and transfers naturally to anomaly detection, where models fitted to nominal data reconstruct anomalous inputs poorly. Benfenati et al.\ introduced masked-autoencoding Transformers on vibration spectrograms as foundation models for SHM, pre-trained on GPU servers and evaluated on proprietary viaduct data, and showed that distilled smaller variants can run on edge nodes \citep{benfenati2024}. The present work adopts that detector family and asks a different question: what architecture around the detector makes inventory-scale deployment possible, and what does a single CPU-sized instance of the detector achieve, against a linear reference, when it is trained and evaluated on public data from structures of different form. Transformer models have also been applied to structural seismic response prediction \citep{zhang2024transformer}.

\subsection{Edge and fog computing for SHM}
Edge computing targets latency, bandwidth and resilience \citep{shi2016}; fog computing adds a regional layer between devices and cloud \citep{bonomi2012}. In SHM, on-node processing achieved 99\% data reduction in a wireless bridge network \citep{hackmann2014}; wireless sensing surveys identify power, synchronization and data management as the persistent constraints \citep{lynch2006,sony2019}. Recent edge SHM systems advance individual capabilities (Table~\ref{tab:edgecompare}). Buckley et al.\ extract features on a microcontroller with LoRa uplink from single-channel data \citep{buckley2021}; Hidalgo-Fort et al.\ sample eight minutes once per day to reach a ten-year battery life, processing after collection \citep{hidalgofort2024}; Mishra et al.\ run binary crack classification on a neural processing unit from static images \citep{mishra2023}; Mart\'in et al.\ propose an edge/fog/cloud architecture, validated on an in-service tunnel, in which damage detection by modal analysis runs in the fog tier rather than on the node \citep{martin2022}; see also \citep{ribeiro2023,zhang2025edge}. Each operates in a periodic or single-modality regime, or is validated on laboratory specimens or private data. The architecture presented here integrates continuous processing, sub-second on-node inference, a channel-agnostic multi-sensor representation and autonomous operation in one edge-deployable design, validated on public data from three in-service bridges.

\begin{table}[t]
\centering\small
\caption{Capabilities of recent edge SHM systems and of the present architecture. \yes\ demonstrated; \pmark\ partial; \no\ not addressed. Entries compiled from the cited papers. $^{a}$~in-service tunnel, not a bridge. $^{b}$~public crack-image training set; test images partly private. $^{c}$~fog tier can run without the cloud; the node itself does not decide.}
\label{tab:edgecompare}
\begin{tabular}{@{}p{0.36\linewidth}ccccc@{}}
\toprule
Capability & \shortstack{Buckley\\ et al.~\cite{buckley2021}} & \shortstack{Hidalgo-Fort\\ et al.~\cite{hidalgofort2024}} & \shortstack{Mishra\\ et al.~\cite{mishra2023}} & \shortstack{Mart\'in\\ et al.~\cite{martin2022}} & \shortstack{This\\ work} \\
\midrule
Continuous (not periodic) processing & \pmark & \no & \no & \pmark & \yes \\
On-node inference, sub-second latency & \pmark & \no & \yes & \no & \yes \\
Multi-channel vibration representation & \no & \pmark & \no & \pmark & \yes \\
Validated on in-service bridge data & \no & \yes & \no & \pmark$^{a}$ & \yes\ (three) \\
Reproducible from public data & \no & \no & \pmark$^{b}$ & \no & \yes \\
Autonomous during network loss & \pmark & \pmark & \pmark & \pmark$^{c}$ & \yes \\
Fleet-scale tiered architecture & \no & \pmark & \no & \yes & \yes \\
\bottomrule
\end{tabular}
\end{table}

\section{Reference Architecture}\label{sec:arch}

\subsection{Design principles}
\begin{figure}[t]
\centering
\resizebox{\linewidth}{!}{% Figure: three-tier reference architecture (edge / fog / cloud)
\begin{tikzpicture}[
  font=\small,
  tier/.style={draw=gray!60, rounded corners=4pt, fill=gray!6, inner sep=8pt},
  box/.style={draw=blue!60!black, fill=white, rounded corners=2pt, align=center, minimum height=8mm, inner sep=3pt, text width=21mm, font=\footnotesize},
  gbox/.style={box, draw=green!50!black},
  pbox/.style={box, draw=violet!70!black},
  lbl/.style={font=\footnotesize\bfseries, text=gray!70!black},
  flow/.style={-{Latex[length=2mm]}, thick, draw=gray!60!black},
  back/.style={-{Latex[length=2mm]}, thick, dashed, draw=violet!70!black},
  node distance=4mm and 8mm]

% ---- edge tier
\node[box] (sens) {Accelerometers\\ 100--200 Hz, multi-channel};
\node[box, right=of sens] (win) {Windowing\\ 60 s per channel};
\node[box, right=of win] (fft) {Hanning + FFT\\ per-structure spectrogram};
\node[gbox, right=of fft] (model) {Detector (PCA or MAE-ViT)\\ reconstruction error};
\node[gbox, right=of model] (thr) {Threshold\\ median $+2.5\sigma$};
\node[box, below=6mm of fft] (buf) {Local store\\ ring buffer};
\node[box, below=6mm of thr] (cb) {Circuit breaker\\ store-and-forward};

% ---- fog tier
\node[box, anchor=north west] (kafka) at ([yshift=-17mm]sens.west |- buf.south) {Message broker\\ (Kafka) durable queue};
\node[box, right=of kafka] (flink) {Stream processing\\ (Flink) dedup, enrich, filter};
\node[box, right=of flink] (regdb) {Regional statistics\\ anomaly rates, latency};

% ---- cloud tier
\node[box, anchor=north west] (ts) at ([yshift=-17mm]sens.west |- kafka.south) {Time-series store\\ long-term retention};
\node[box, right=of ts] (an) {Fleet analytics\\ nightly batch, trends};
\node[pbox, right=of an] (ml) {Model lifecycle\\ retrain, staged rollout};
\node[box, right=of ml] (al) {Alerting\\ severity routing};

% ---- tier backgrounds (behind the boxes)
\begin{scope}[on background layer]
\node[tier, fit=(sens)(win)(fft)(model)(thr)(buf)(cb)] (edge) {};
\node[tier, fit=(kafka)(flink)(regdb)] (fog) {};
\node[tier, fit=(ts)(an)(ml)(al)] (cloud) {};
\end{scope}
\node[lbl, anchor=south west, inner sep=1pt] at (edge.north west) {EDGE TIER -- one node per structure, no network in the processing path};
\node[lbl, anchor=south west, inner sep=1pt] at (fog.north west) {FOG TIER -- regional hub, 100--500 structures};
\node[lbl, anchor=south west, inner sep=1pt] at (cloud.north west) {CLOUD TIER -- fleet-wide};

% ---- flows
\draw[flow] (sens) -- (win);
\draw[flow] (win) -- (fft);
\draw[flow] (fft) -- (model);
\draw[flow] (model) -- (thr);
\draw[flow] (thr) -- (cb);
\draw[flow] (model) |- (buf);
\draw[flow] (cb.south) -- ++(0,-6mm) -| (flink.north) node[pos=.25, above, font=\scriptsize] {MQTT: health summary + anomaly events ($\approx$1 KB)};
\draw[flow] (kafka) -- (flink);
\draw[flow] (flink) -- (regdb);
\draw[flow] (regdb.south) -- ++(0,-5mm) -| (an.north);
\draw[flow] (an) -- (ml);
\draw[flow] (ml) -- (al);
\draw[flow] (an) -- (ts);
\draw[back] (ml.north) -- ++(0,4mm) -| ([xshift=8mm]edge.east |- thr) -- (thr.east);
\path ([xshift=8mm]edge.east |- thr) -- ([xshift=8mm]edge.east |- ml) node[midway, rotate=90, above, font=\scriptsize, text=violet!70!black] {model updates};
\end{tikzpicture}}
\caption{Three-tier reference architecture. Solid arrows are the always-on data path; the dashed arrow is model distribution. The edge tier runs without any dependency to the right of it.}
\label{fig:arch}
\end{figure}
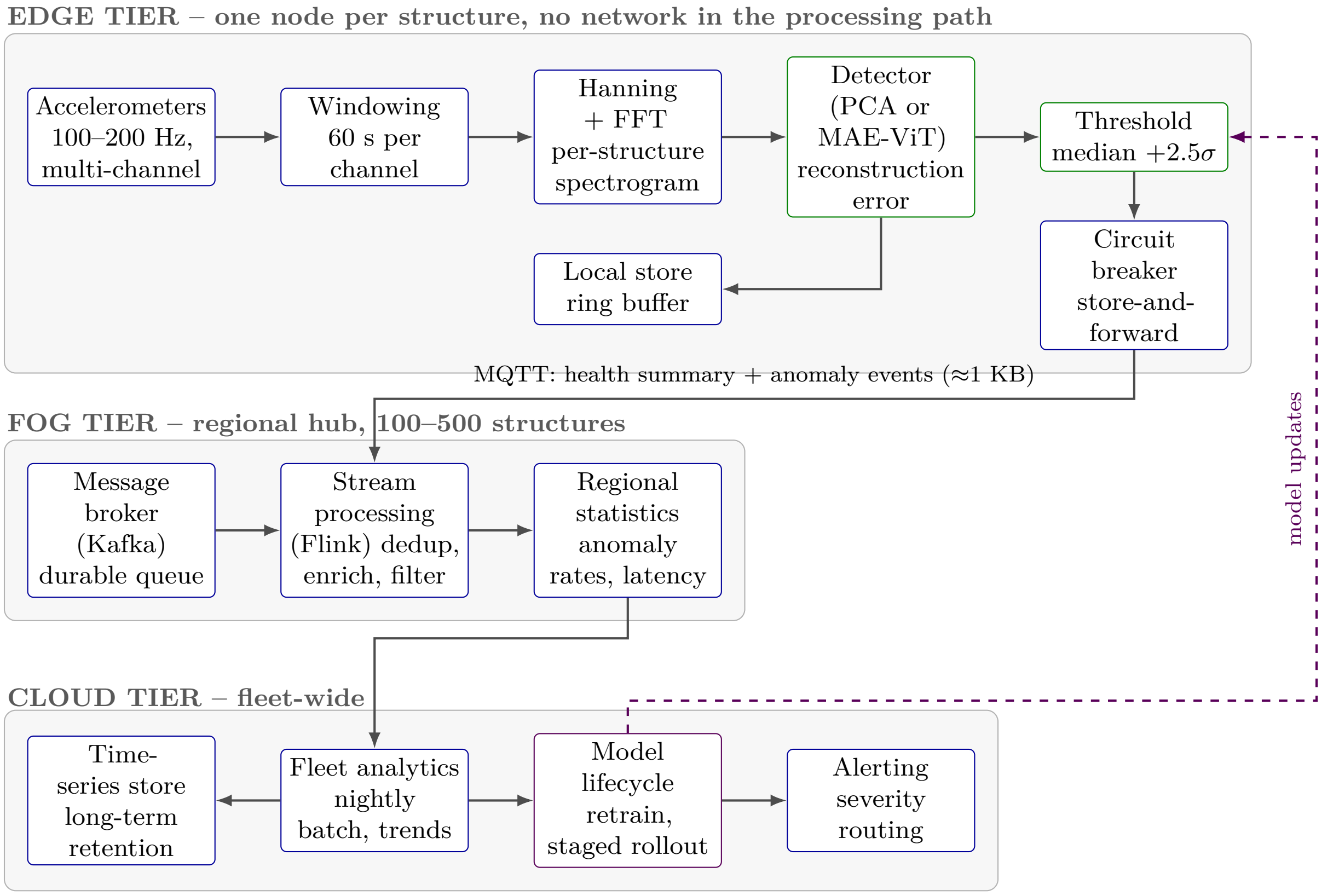

Four principles govern the design. \emph{Process at the source:} every window is analyzed at the structure, so decision latency is bounded by on-node compute rather than by network transport. \emph{Transmit summaries, not waveforms:} the uplink carries health indicators and anomaly events; raw windows stay in a local ring buffer and are served on request. \emph{Run on commodity hardware:} the edge node is specified at 4\,GB of memory and a general-purpose CPU, which places per-structure cost in the range of a single-board computer. \emph{Stay autonomous:} no network call sits in the processing path; connectivity loss degrades reporting latency, never analysis. Figure~\ref{fig:arch} shows the three tiers; only the edge tier is prescribed in detail, the fog and cloud tiers are integration points with defined contracts.

\subsection{Edge tier}
A persistent worker process loads the detector once and then consumes windows from a local queue, amortizing initialization over thousands of inferences. Each window passes through preprocessing (resampling, windowing, Hanning taper, FFT, spectrogram, normalization), scoring, thresholding and publication. Lazy generators decouple producer and consumer rates and provide natural backpressure. A circuit breaker monitors the fog link; on failure it switches to store-and-forward, buffering results on local storage and replaying them in order on reconnection \citep{nygard2018}. The target class of device is a 4\,GB single-board computer with a 5--10\,W power budget suitable for solar supply at remote sites. The detector is a replaceable component behind a fixed interface (spectrogram in, score out); Section~\ref{sec:results} evaluates two.

\subsection{Fog tier}
A regional hub aggregates 100--500 structures. A durable message broker (Apache Kafka) provides replay and horizontal scaling; a stream processor (Apache Flink) deduplicates messages replayed after outages, validates and enriches records, computes regional statistics (anomaly rates, latency percentiles, trends) and separates high-priority anomaly events, forwarded immediately, from routine summaries batched at one-minute intervals. The tier reduces cloud ingestion by roughly a further 10:1 and keeps dashboard queries regional. Twenty hubs cover the continental United States in the cost model of Section~\ref{sec:cost}.

\subsection{Cloud tier}
The cloud tier holds long-term time-series storage with lifecycle tiering, nightly fleet analytics that correlate anomalies across structures and trend degradation, periodic model retraining, and alert routing by severity with six-hour deduplication. Updated models are distributed through a staged rollout (10\%, 50\%, 100\% of nodes) with automatic rollback on metric regression \citep{kleppmann2017}.

\subsection{Message contract and data reduction}
Each edge message carries structure and sensor identifiers, timestamp, reconstruction error, binary decision, confidence, processing latency and optional context (temperature, traffic count) in about 200 bytes of compressed JSON, one message per channel per window. For ten channels sampled at 100\,Hz the raw stream is about 4\,kB/s and the message stream about 33\,B/s, a reduction of roughly 120:1 at the edge and a further 10:1 after fog batching; for the 12-channel, 825.8\,Hz instrumentation of KW51 the edge reduction is about 1000:1.

\section{Edge Detection Pipeline}\label{sec:edge}

\subsection{Signal representation}\label{sec:signal}
The front end turns one channel of one window into a $100\times200$ image of $\log_{10}$ power spectral density (100 time frames by 200 frequency bins) and is the point at which structural dynamics enters the pipeline. Its parameters are set per structure from the documented behaviour of the bridge (Table~\ref{tab:frontend}), not learned.

\textbf{Window and resolution.} Windows are 60\,s. For KW51 and Chillon the channel is resampled to 100\,Hz with a polyphase anti-aliasing filter; Hanning frames of 20\,s stepped by 0.4\,s give 100 frames, and a 2{,}000-point FFT gives 0.05\,Hz bins, of which the 200 from 0 to 9.95\,Hz are kept. The resolution follows from the documented effect: the retrofit of KW51 shifted its fourteen tracked modes, which lie at 0.5--7.1\,Hz, by 0.3--2.7\% \citep{maes2021}, that is by 0.02--0.19\,Hz, so a bin width of 0.05\,Hz is needed for the shift to move energy across pixels. For V\"anersborg the fracture is a local event in one member excited during bridge openings, with no documented effect on the low modes; the native 200\,Hz record is kept, frames are 2\,s stepped by 0.585\,s and a 400-point FFT gives 0.5\,Hz bins from 0 to 99.5\,Hz.

\textbf{Channels.} Each sensor channel is an independent window, so the detector is channel-agnostic and a structure contributes as many windows per minute as it has channels. KW51 is restricted to the six deck accelerometers present in both monitoring years, because the arch sensors exist only in the reference year and would let a detector identify the condition from the sensor set rather than from the structure.

\textbf{Normalization.} The mean $\log_{10}$ PSD of each window is subtracted before the image is scaled with the global minimum and maximum of the training windows. Without the first step the reconstruction error of either detector tracks the absolute vibration level of the window ($r=0.89$ on KW51 in a preliminary run), which varies with traffic and temperature and is not a property of the structure.

The first configuration tried, a single front end for all three structures with 11.9\,s windows, 0.1\,Hz bins and absolute level retained, detected neither documented change. The diagnostics that led to the present front ends (bin width against the documented modal shift, frequency band of the V\"anersborg signature, error--level correlation, channel-set asymmetry) are released with the code; the band, window and resolution are configuration, so a structure class outside these settings is accommodated by configuration rather than redesign.

\begin{table}[t]
\centering\small
\caption{Per-structure front end. Every configuration yields a $100\times200$ log-PSD image per channel per window.}
\label{tab:frontend}
\begin{tabular}{@{}lccc@{}}
\toprule
 & KW51 & V\"anersborg & Chillon \\
\midrule
Channels used & 6 deck of 12 & 5 accelerometers & 11 \\
Sampling (Hz) & 100 (from 825.8) & 200 (native) & 100 \\
Window (s) & 59.6 & 59.9 & 59.6 \\
Frame / step (s) & 20 / 0.4 & 2 / 0.585 & 20 / 0.4 \\
Bin width (Hz) & 0.05 & 0.5 & 0.05 \\
Band (Hz) & 0--9.95 & 0--99.5 & 0--9.95 \\
Level normalization & per window & per window & per window \\
\bottomrule
\end{tabular}
\end{table}

\subsection{Detector: linear reconstruction error}\label{sec:pca}
The deployed detector is principal component analysis of the normalized image, fitted on the training windows of the structure and retaining the components that explain 95\% of their variance (\pcaKKW\ for KW51, \pcaKVB\ for V\"anersborg, \pcaKCH\ for Chillon). The anomaly score $e$ of a window is the mean squared residual outside that subspace. The detector has no hyperparameters beyond the variance fraction, fits in seconds on the node, is deterministic, and its cost per window is one matrix product.

\subsection{Learned alternative: masked-autoencoding Vision Transformer}\label{sec:vit}
\begin{figure}[t]
\centering
\resizebox{\linewidth}{!}{% Figure: MAE-ViT anomaly detector scaled for edge inference
\begin{tikzpicture}[
  font=\footnotesize,
  blk/.style={draw=blue!60!black, fill=blue!4, rounded corners=2pt, align=center, minimum height=9mm, inner sep=4pt},
  dec/.style={blk, draw=orange!80!black, fill=orange!6},
  outb/.style={blk, draw=green!50!black, fill=green!5},
  flow/.style={-{Latex[length=2mm]}, thick, draw=gray!60!black},
  node distance=5mm and 6mm]

\node[blk] (in) {Spectrogram\\ $100\times200$ log-PSD};
\node[blk, right=of in] (patch) {Patch embedding\\ $5\times10$ px, 400 patches\\ $\to$ 768-d};
\node[blk, right=of patch] (enc) {Encoder\\ 3 layers, 12 heads\\ 21.3 M params};
\node[dec, right=of enc] (mask) {Random mask\\ 80\% of patches\\ (training and scoring)};
\node[dec, right=of mask] (decd) {Decoder\\ 4 layers, 512-d, 16 heads\\ 13.0 M params};
\node[outb, below=of decd] (rec) {Reconstruction\\ $\hat{X}$};
\node[outb, left=of rec] (err) {Masked-patch error\\ $e=\frac{1}{4}\sum_{m=1}^{4}\|M_m(X-\hat{X}_m)\|_2^2/|M_m|$};
\node[outb, left=of err] (dec2) {Anomaly if $e > \tau$\\ $\tau=\mathrm{med}(e_{\mathrm{normal}})+2.5\,\mathrm{sd}$};

\draw[flow] (in) -- (patch);
\draw[flow] (patch) -- (enc);
\draw[flow] (enc) -- (mask);
\draw[flow] (mask) -- (decd);
\draw[flow] (decd) -- (rec);
\draw[flow] (rec) -- (err);
\draw[flow] (err) -- (dec2);
\draw[flow, dashed] (in.south) |- (err.west) node[pos=.65, below, font=\scriptsize] {original $X$};

\node[below=2mm of dec2.south west, anchor=north west, font=\scriptsize, text=gray!60!black, align=left]
 {Training: reference-condition windows only, MSE on masked patches. Scoring: four masked passes, same error, averaged.\\
  Footprint: 34.3 M parameters, 139 MB on disk, under 1 GB resident incl.\ runtime; CPU-only. Scored unmasked the same weights fail (Section 6.5).};
\end{tikzpicture}}
\caption{Masked-autoencoding Vision Transformer scaled for CPU-only edge inference, scored on masked patches as in training.}
\label{fig:model}
\end{figure}
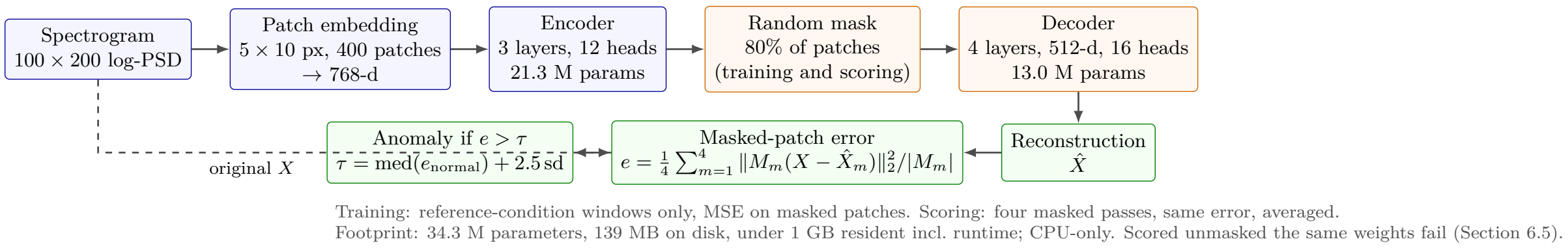

The learned detector is a masked-autoencoding Vision Transformer (Figure~\ref{fig:model}). The image is split into 400 patches of $5\times10$ pixels, each projected to a 768-dimensional embedding with fixed two-dimensional sine--cosine position encoding. The encoder has 3 transformer layers (12 heads, 3{,}072-wide feed-forward, pre-layer-norm, residual connections), 21.3 million parameters; the decoder has 4 layers at 512 dimensions and 16 heads, 13.0 million; \modelParams\ million in total. Training masks 80\% of the patches and minimizes mean squared reconstruction error on the masked patches of reference-condition windows only. The depth reduction from the 12-layer standard configuration is a deliberate trade for the edge budget.

The anomaly score is computed in the same regime: the window is passed four times with independent random 80\% masks and $e$ is the mean squared error on the masked patches, averaged over the four passes. The alternative of encoding and decoding the full, unmasked image and taking the error over all patches, which costs one pass instead of four, is also reported in Section~\ref{sec:cmp} because it is the obvious choice and it fails: a model that has only ever reconstructed from 20\% of the patches produces an error on the full image that is an order of magnitude larger than its training loss and that ranks the two conditions in the wrong order.

\subsection{Threshold calibration}\label{sec:thr}
Temperature, traffic and wind change natural frequencies without damage. The decision threshold is therefore set per structure and per detector from the distribution of $e$ on a calibration set of reference-condition windows disjoint from training, at the median plus 2.5 standard deviations, an operating point chosen so that alerts indicate departure from established reference behaviour rather than operational fluctuation. Every new installation runs a calibration period before alerting is enabled, and the threshold is refreshed on a schedule to follow slow environmental drift. The multiplier is an explicit parameter: Section~\ref{sec:sweep} reports the full sensitivity--specificity trade-off as it is varied, so that an owner who values detection over false alarms can choose a different point with known consequences.

\subsection{Streaming runtime and latency accounting}
Per-window latency is measured from receipt of the last sample to publication of the decision, at batch size one, and decomposed into spectrogram, normalization, scoring and threshold stages. A node that falls behind the data rate accumulates a backlog that defeats continuous monitoring even though structural change itself evolves over hours to years; with one window per channel per minute, a node serving twelve channels must decide in well under five seconds per window, and the sub-second figures of Section~\ref{sec:latency} leave room for denser instrumentation or a slower processor.

\section{Data and Protocol}\label{sec:setup}

\subsection{Three public in-service bridges}
Validation uses three openly licensed monitoring datasets from in-service bridges, chosen so that the structural forms differ and so that two of them contain a documented, independently verified change of condition (Table~\ref{tab:data}).

\textbf{KW51, steel bowstring railway bridge (Leuven, Belgium).} Monitored for fifteen months by KU Leuven with twelve accelerometers on deck and arches at 825.8\,Hz, in six five-minute ambient records per day \citep{maes2021,maes2020data}. The connections of the diagonals to the deck and arches were strengthened between 15 May and 27 September 2019 after damage had been observed. The reference condition is January 2020, after the retrofit; the changed condition is January 2019, before it, with the deficient connections in place. Using the same calendar month on both sides keeps temperature from standing in for the structural change. The documented effect is a shift of the tracked natural frequencies by 0.3--2.7\%.

\textbf{V\"anersborg, riveted steel bascule railway bridge (Sweden).} Built 1914--1916 and monitored by KTH and IoTBridge with five accelerometers and sixteen strain gauges at 200\,Hz during bridge-opening events \citep{leander2023,leander2023data}. During the opening of 9 March 2023 at 23:45 a crack propagated in a member of the counterweight truss. Only the accelerometers are used here. The reference condition is the pre-fracture openings; the changed condition is the post-fracture openings; the opening during which the crack propagated is excluded from both.

\textbf{Chillon, prestressed concrete box-girder viaduct (Switzerland).} Strengthened with ultra-high-performance fibre-reinforced concrete and monitored by ETH Z\"urich with eleven vertical accelerometers at 100\,Hz in ten-minute records with air temperature \citep{martinsanz2020,martinsanz2019data}. The structure was in normal operation throughout; this dataset measures the false-alarm rate of a calibrated detector under a season of temperature and traffic variation (\tempMinCH--\tempMaxCH\,\si{\celsius}). Records spanning 26 March to 8 June 2017 are used.

\begin{table}[t]
\centering\small
\caption{Public datasets. Windows are 60\,s per channel; test counts are the windows that enter the reported metrics.}
\label{tab:data}
\footnotesize
\begin{tabular}{@{}p{0.13\linewidth}p{0.24\linewidth}p{0.19\linewidth}p{0.19\linewidth}rr@{}}
\toprule
Structure & Form, sensors & Reference condition & Changed condition & \multicolumn{2}{c}{Test windows} \\
 & & & & ref. & changed \\
\midrule
KW51 \citep{maes2020data} & Steel bowstring railway bridge; 12 acc., 825.8\,Hz & Jan 2020, after connection retrofit & Jan 2019, deficient diagonal connections & \nNormalKW & \nAnomalyKW \\
V\"anersborg \citep{leander2023data} & Riveted steel bascule railway bridge; 5 acc., 200\,Hz & openings before 9 Mar 2023 & openings after verified fracture & \nNormalVB & \nAnomalyVB \\
Chillon \citep{martinsanz2019data} & Prestressed concrete box-girder viaduct; 11 acc., 100\,Hz & Mar--Jun 2017, normal operation, \tempMinCH--\tempMaxCH\,\si{\celsius} & none & \nNormalCH & -- \\
\bottomrule
\end{tabular}
\end{table}

\subsection{Splits, training and calibration}
Within each structure, reference-condition source files (one five- or ten-minute record, or one bridge opening) are assigned by a seeded draw to training (60\%), calibration (20\%) and test (20\%); changed-condition files are test only. Whole files stay together, so no window of a test record is seen in training or calibration. Both detectors are fitted per structure on the same training windows: \nTrainKW\ for KW51, \nTrainVB\ for V\"anersborg and \nTrainCH\ for Chillon (a cap of 2{,}500 drawn from its larger training split). PCA is fitted once per structure. The Transformer is trained once per structure with AdamW (peak learning rate $10^{-3}$, cosine decay, two warm-up epochs, batch 32) for at least 1{,}500 optimizer steps, which is \epochsKW, \epochsVB\ and \epochsCH\ epochs respectively, on a 32-core CPU; final masked reconstruction losses are \finalLossKW, \finalLossVB\ and \finalLossCH\ (Appendix~\ref{app:train}). Thresholds (Section~\ref{sec:thr}) are computed per structure and per detector on the calibration windows.

\subsection{Metrics}
Detection is reported as the area under the receiver-operating curve (AUC) over all thresholds, and as sensitivity, specificity, precision and F1 at the calibrated threshold with the confusion matrix and 95\% Wilson intervals for the headline proportions. Because an operator runs a detector at a fixed false-alarm budget, sensitivity at 1\% and 5\% false-alarm rate on the test windows is also given. Receiver-operating curves and the metrics as a function of the threshold multiplier are reported for both detectors.

\subsection{Edge latency protocol}
Latency is measured inside a container limited to 4 virtual CPUs and 4\,GB of memory, with no accelerator, running the same Python environment as the edge image (\texttt{torch} \torchVersion, CPU build, for the Transformer), at batch size one, on \latn\ raw test windows drawn from all three structures. Each measurement spans the complete pipeline from the raw window to the published decision and records the time of each stage. The Transformer is timed on one unmasked pass, the budget-setting case for the hardware; its masked score costs four encoder passes on 20\% of the tokens plus four decoder passes.

\section{Results}\label{sec:results}

\subsection{Detection of a documented change of condition}
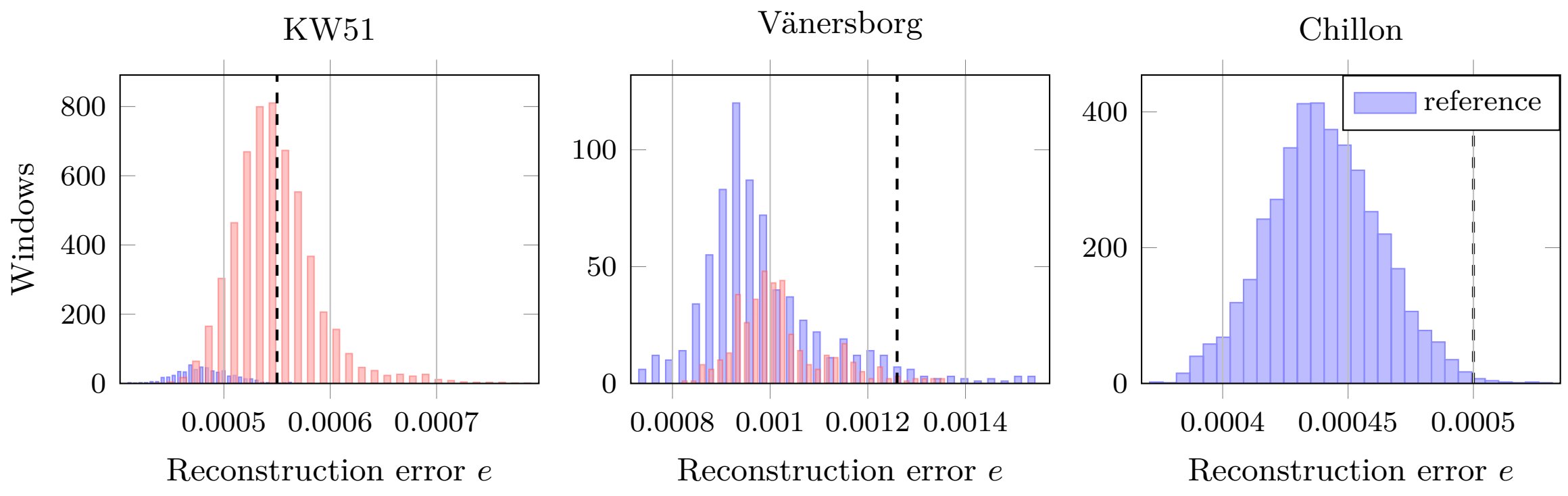
\begin{figure}[t]
\centering
\begin{tikzpicture}
\begin{groupplot}[group style={group size=3 by 1, horizontal sep=9mm}, width=.355\linewidth, height=4.6cm, ybar interval, ymin=0,
  area style, enlarge x limits=0.02, x tick label as interval=false, scaled x ticks=false,
  xticklabel style={/pgf/number format/fixed, /pgf/number format/precision=5}, xlabel={Reconstruction error $e$}]
\nextgroupplot[xtick={\errTicksKW}, title={KW51}, ylabel={Windows}]
\addplot+[hist={bins=30}, normalbar] table[y=err_pca, col sep=comma] {generated/kw51_err_normal.csv};
\addplot+[hist={bins=30}, anomalybar] table[y=err_pca, col sep=comma] {generated/kw51_err_anomaly.csv};
\draw[dashed,thick] ({axis cs:\tauKW,0}|-{rel axis cs:0,0}) -- ({axis cs:\tauKW,0}|-{rel axis cs:0,1});
\nextgroupplot[xtick={\errTicksVB}, title={V\"anersborg}]
\addplot+[hist={bins=30}, normalbar] table[y=err_pca, col sep=comma] {generated/vanersborg_err_normal.csv};
\addplot+[hist={bins=30}, anomalybar] table[y=err_pca, col sep=comma] {generated/vanersborg_err_anomaly.csv};
\draw[dashed,thick] ({axis cs:\tauVB,0}|-{rel axis cs:0,0}) -- ({axis cs:\tauVB,0}|-{rel axis cs:0,1});
\nextgroupplot[xtick={\errTicksCH}, title={Chillon}, legend style={at={(1.0,1.0)}, anchor=north east, font=\scriptsize}]
\addplot+[hist={bins=30}, normalbar] table[y=err_pca, col sep=comma] {generated/chillon_err_normal.csv};
\draw[dashed,thick] ({axis cs:\tauCH,0}|-{rel axis cs:0,0}) -- ({axis cs:\tauCH,0}|-{rel axis cs:0,1});
\legend{reference}
\end{groupplot}
\datawatermark
\end{tikzpicture}
\caption{Error distributions of the deployed detector on the test windows (blue: reference condition; red: changed condition) and the calibrated threshold $\tau$ (dashed) of each structure.}
\label{fig:err}
\end{figure}

\begin{figure}[t]
\centering
\begin{tikzpicture}
\begin{groupplot}[group style={group size=2 by 1, horizontal sep=12mm}, width=.46\linewidth, height=5.4cm,
  xmin=0, xmax=1, ymin=0, ymax=1, xlabel={False-alarm rate ($1-$specificity)}, grid=major, grid style={gray!20}]
\nextgroupplot[title={KW51}, ylabel={Sensitivity}, legend pos=south east, legend style={font=\scriptsize}]
\addplot[pca] table[x=fpr, y=tpr, col sep=comma] {generated/kw51_roc_pca.csv};
\addplot[vit] table[x=fpr, y=tpr, col sep=comma] {generated/kw51_roc_vit.csv};
\addplot[unm] table[x=fpr, y=tpr, col sep=comma] {generated/kw51_roc_vit_unmasked.csv};
\addplot[only marks, mark=*, mark size=2pt, cPca] coordinates {(\fpeval{1-\specKW/100}, \fpeval{\sensKW/100})};
\addplot[gray!50, thin] coordinates {(0,0) (1,1)};
\legend{PCA (deployed), MAE-ViT masked score, MAE-ViT unmasked score, operating point}
\nextgroupplot[title={V\"anersborg}]
\addplot[pca] table[x=fpr, y=tpr, col sep=comma] {generated/vanersborg_roc_pca.csv};
\addplot[vit] table[x=fpr, y=tpr, col sep=comma] {generated/vanersborg_roc_vit.csv};
\addplot[unm] table[x=fpr, y=tpr, col sep=comma] {generated/vanersborg_roc_vit_unmasked.csv};
\addplot[only marks, mark=*, mark size=2pt, cPca] coordinates {(\fpeval{1-\specVB/100}, \fpeval{\sensVB/100})};
\addplot[gray!50, thin] coordinates {(0,0) (1,1)};
\end{groupplot}
\datawatermark
\end{tikzpicture}
\caption{Receiver-operating curves over all thresholds for the deployed detector and for the Transformer under both scoring regimes, with the calibrated operating point of the deployed detector marked.}
\label{fig:roc}
\end{figure}
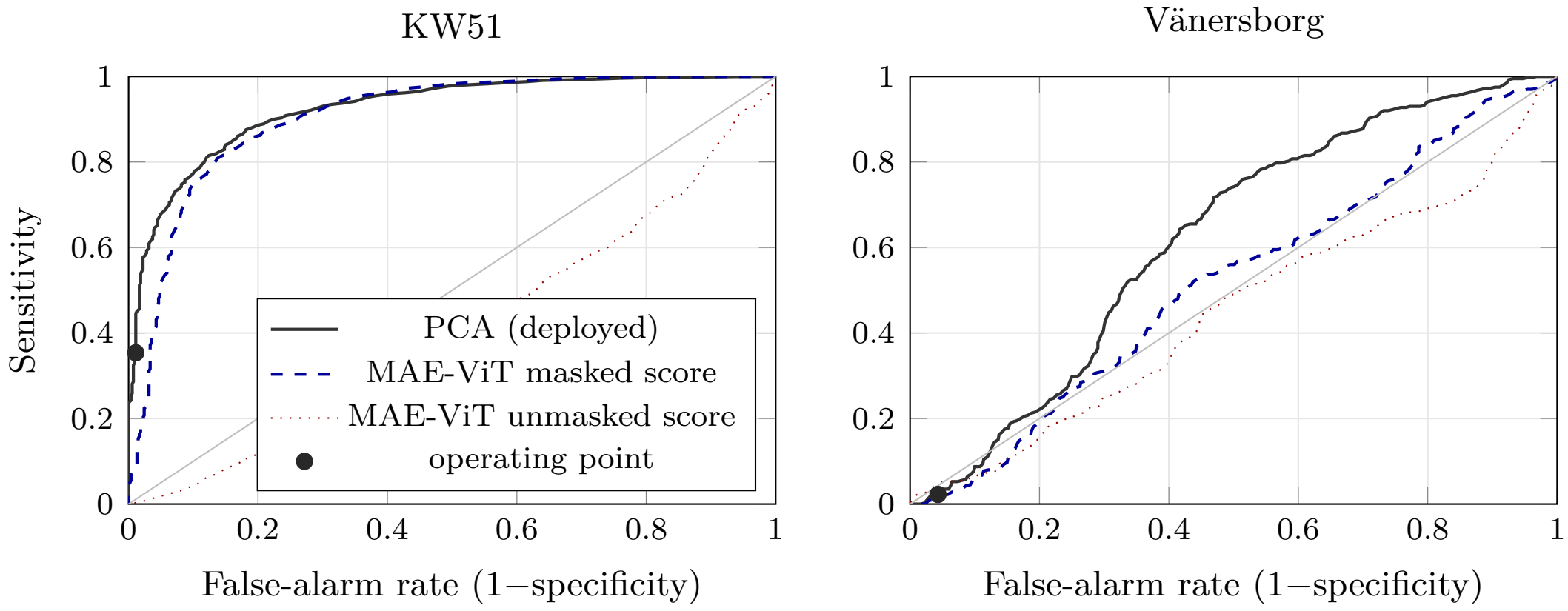

Table~\ref{tab:cls} gives the metrics of the deployed detector on the two structures with a documented change, with the confusion matrices in Table~\ref{tab:conf}. On KW51 the detector separates the two conditions with AUC \aucKW\ and, at the calibrated threshold $\tau=\tauKW$, flags \sensKW\% of the pre-retrofit windows (95\% CI $\pm$\ciSensKW\,pp) while classifying \specKW\% of the post-retrofit windows as reference (CI $\pm$\ciSpecKW\,pp); at a 1\% false-alarm budget it flags \sensAtOneKW\% of the changed windows, at 5\% \sensAtFiveKW\%. Each changed window is one minute of one channel, so a detection rate of over a third at a 1\% false-alarm rate means that a deficient connection declares itself within minutes on a bridge whose six deck channels each produce a decision every minute. On V\"anersborg the detector does not separate the conditions: AUC \aucVB, \sensVB\% sensitivity (CI $\pm$\ciSensVB\,pp) at \specVB\% specificity. Section~\ref{sec:vb} discusses what this bounds. Figure~\ref{fig:err} shows the two error populations of each structure against its threshold; Figure~\ref{fig:roc} shows the receiver-operating curves of both detectors.

\begin{table}[t]
\centering\small
\caption{Deployed detector (PCA) at the calibrated operating point (median $+2.5$\,sd of calibration error), per-structure front end and threshold.}
\label{tab:cls}
\begin{tabular}{@{}lS[table-format=2.3]S[table-format=2.3]S[table-format=2.3]@{}}
\toprule
 & {KW51} & {V\"anersborg} & {Chillon} \\
\midrule
AUC & \aucKW & \aucVB & {--} \\
Sensitivity (\%) & \sensKW & \sensVB & {--} \\
Specificity (\%) & \specKW & \specVB & \specCH \\
Precision (\%) & \precKW & \precVB & {--} \\
F1 score (\%) & \fKW & \fVB & {--} \\
Sensitivity at 1\% false alarms (\%) & \sensAtOneKW & \sensAtOneVB & {--} \\
Sensitivity at 5\% false alarms (\%) & \sensAtFiveKW & \sensAtFiveVB & {--} \\
\midrule
Threshold $\tau$ & \tauKW & \tauVB & \tauCH \\
PCA components & {\pcaKKW} & {\pcaKVB} & {\pcaKCH} \\
Test windows & {\nKW} & {\nVB} & {\nCH} \\
\bottomrule
\end{tabular}
\end{table}

\begin{table}[t]
\centering\small
\caption{Confusion matrices of the deployed detector (windows).}
\label{tab:conf}
\footnotesize
\begin{tabular}{@{}lcccccc@{}}
\toprule
 & \multicolumn{2}{c}{KW51} & \multicolumn{2}{c}{V\"anersborg} & \multicolumn{2}{c}{Chillon} \\
 & pred.\ ref. & pred.\ changed & pred.\ ref. & pred.\ changed & pred.\ ref. & pred.\ changed \\
\midrule
Reference condition & \tnKW & \fpKW & \tnVB & \fpVB & \tnCH & \fpCH \\
Changed condition & \fnKW & \tpKW & \fnVB & \tpVB & -- & -- \\
\bottomrule
\end{tabular}
\end{table}

\subsection{Error over time}
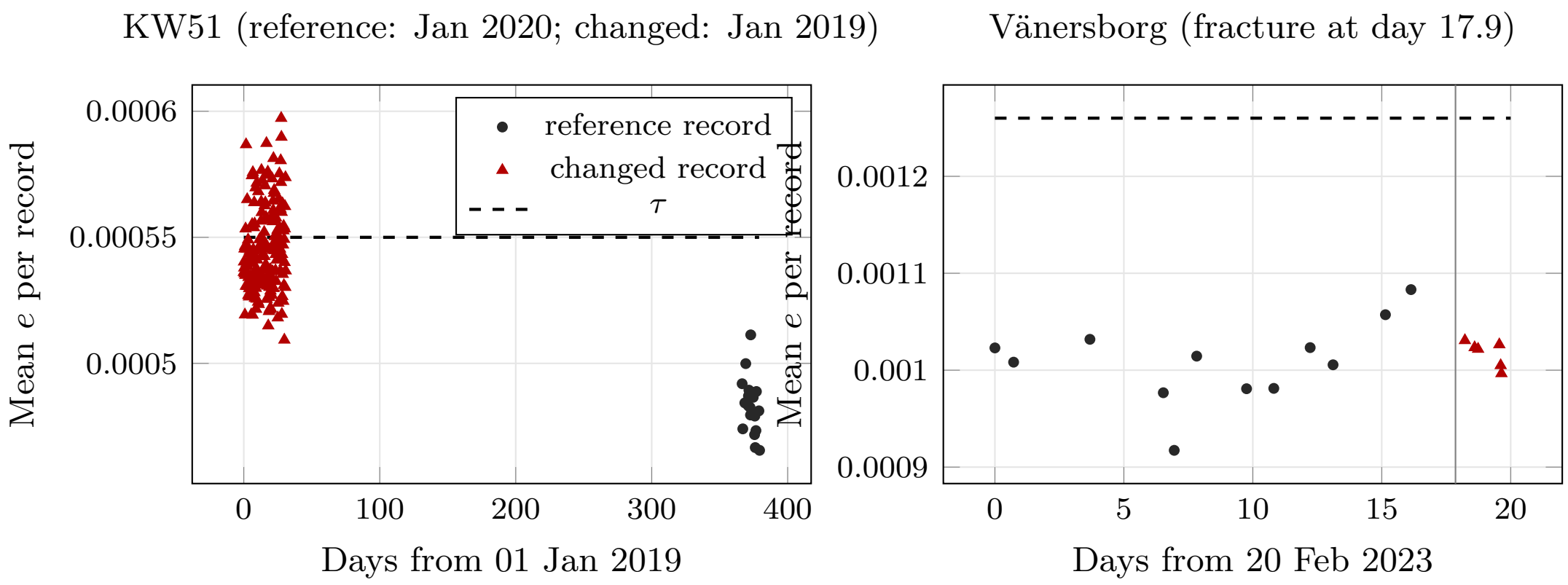
\begin{figure}[t]
\centering
\begin{tikzpicture}
\begin{groupplot}[group style={group size=2 by 1, horizontal sep=12mm}, width=.45\linewidth, height=5.2cm,
  ylabel={Mean $e$ per record}, grid=major, grid style={gray!20}, legend style={font=\scriptsize},
  yticklabel style={/pgf/number format/fixed, /pgf/number format/precision=5}, scaled y ticks=false]
\nextgroupplot[title={KW51 (reference: Jan 2020; changed: Jan 2019)}, xlabel={Days from \dayStartKW}, legend pos=north east]
\addplot[only marks, mark=*, mark size=1.2pt, cPca] table[x=day, y=pca_mean, col sep=comma] {generated/kw51_files_normal.csv};
\addplot[only marks, mark=triangle*, mark size=1.5pt, red!70!black] table[x=day, y=pca_mean, col sep=comma] {generated/kw51_files_anomaly.csv};
\addplot[dashed, thick] coordinates {(0,\tauKW) (\daySpanKW,\tauKW)};
\legend{reference record, changed record, $\tau$}
\nextgroupplot[title={V\"anersborg (fracture at day 17.9)}, xlabel={Days from \dayStartVB}]
\addplot[only marks, mark=*, mark size=1.2pt, cPca] table[x=day, y=pca_mean, col sep=comma] {generated/vanersborg_files_normal.csv};
\addplot[only marks, mark=triangle*, mark size=1.5pt, red!70!black] table[x=day, y=pca_mean, col sep=comma] {generated/vanersborg_files_anomaly.csv};
\addplot[dashed, thick] coordinates {(0,\tauVB) (\daySpanVB,\tauVB)};
\draw[gray, thin] ({axis cs:17.86,0}|-{rel axis cs:0,0}) -- ({axis cs:17.86,0}|-{rel axis cs:0,1});
\end{groupplot}
\datawatermark
\end{tikzpicture}
\caption{Mean error of the deployed detector for each test record (five-minute ambient record for KW51, bridge opening for V\"anersborg) against time, with the calibrated threshold. KW51 records are plotted on a common day axis for the two Januaries.}
\label{fig:time}
\end{figure}

Figure~\ref{fig:time} plots the mean error of every test record against time. On KW51 the two populations are a year apart and the retrofit lies between them. At record level they do not overlap: every post-retrofit test record has a mean error at or below \recRefMaxKW\ and every pre-retrofit record at or above \recChgMinKW. The calibrated threshold is set for single-window decisions and sits inside the pre-retrofit population, so \recChgAboveKW\ of the \recChgKW\ pre-retrofit records exceed it on average and none of the \recRefKW\ post-retrofit records do; a fog-tier rule on the record mean would separate the two conditions completely on this data. On V\"anersborg the fracture falls inside a three-week record and the post-fracture openings do not separate from the preceding ones. The per-record view is the quantity the fog tier aggregates: a node reports per-window decisions, the hub trends them.

\subsection{False alarms under environmental variability}\label{sec:chillon}
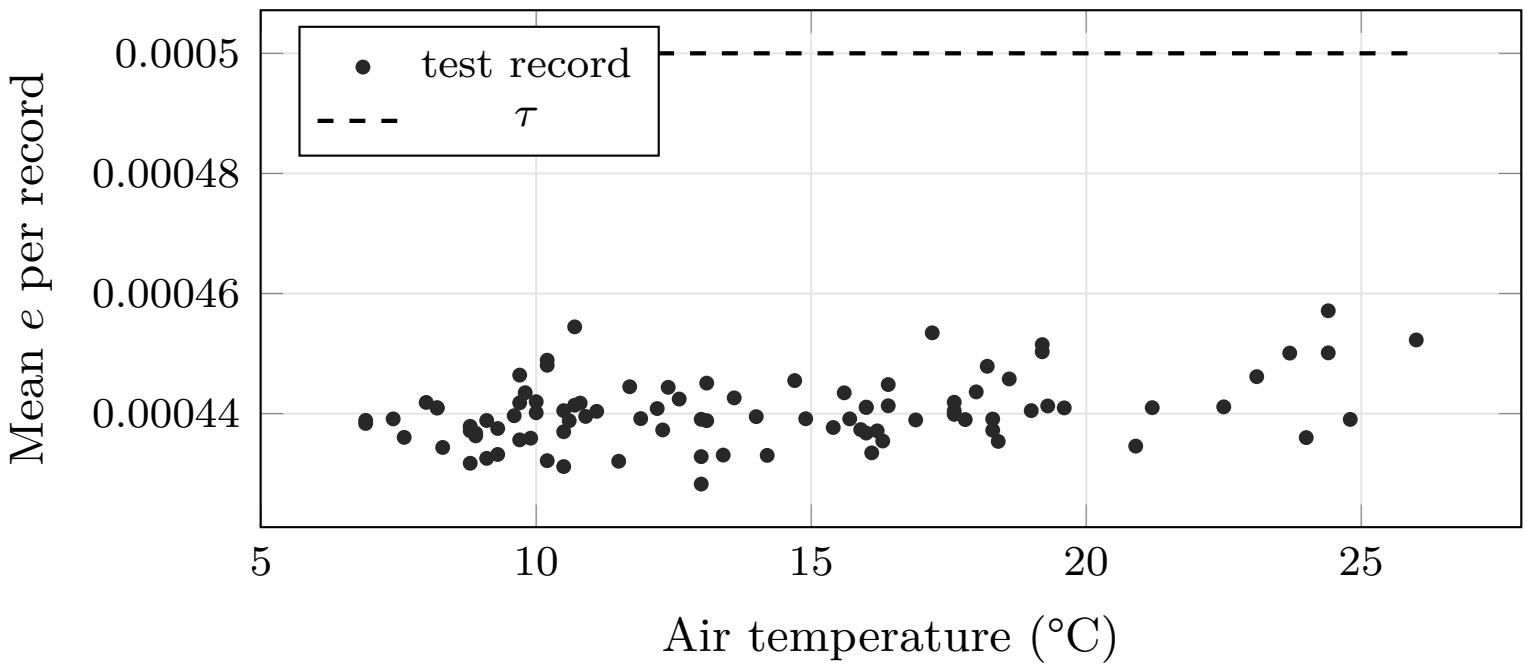
\begin{figure}[t]
\centering
\begin{tikzpicture}
\begin{axis}[width=.62\linewidth, height=5cm, xlabel={Air temperature (\si{\celsius})}, ylabel={Mean $e$ per record}, grid=major, grid style={gray!20}, legend pos=north west, legend style={font=\scriptsize},
  yticklabel style={/pgf/number format/fixed, /pgf/number format/precision=5}, scaled y ticks=false]
\addplot[only marks, mark=*, mark size=1.2pt, cPca] table[x=temp, y=pca_mean, col sep=comma] {generated/chillon_files_normal.csv};
\addplot[dashed, thick, domain=\tempMinCH:\tempMaxCH] {\tauCH};
\legend{test record, $\tau$}
\end{axis}
\datawatermark
\end{tikzpicture}
\caption{Chillon viaduct: mean error of the deployed detector for each ten-minute test record against air temperature, with the calibrated threshold.}
\label{fig:chillon}
\end{figure}

Chillon carries no change of condition; it measures what a calibrated detector does on a healthy structure through spring. Specificity of the deployed detector on \nNormalCH\ test windows is \specCH\% (\fpCH\ false alarms, CI $\pm$\ciSpecCH\,pp) over records spanning \tempMinCH--\tempMaxCH\,\si{\celsius}; the Transformer gives \specViTCH\%. Figure~\ref{fig:chillon} plots the per-record mean error against the air temperature recorded with each file: the slope of this relation is the environmental sensitivity that the calibration period absorbs and that the scheduled threshold refresh of Section~\ref{sec:thr} tracks.

\subsection{Operating point}\label{sec:sweep}
\begin{figure}[t]
\centering
\begin{tikzpicture}
\begin{groupplot}[group style={group size=3 by 1, horizontal sep=9mm}, width=.37\linewidth, height=4.8cm,
  xmin=0, xmax=6, ymin=0, ymax=1.02, xlabel={Threshold multiplier $k$}, grid=major, grid style={gray!20}, legend style={font=\scriptsize}]
\nextgroupplot[title={KW51}, ylabel={Rate}, legend pos=south east]
\addplot[pca] table[x=k, y=spec_pca, col sep=comma] {generated/kw51_sweep.csv};
\addplot[pca, dotted] table[x=k, y=sens_pca, col sep=comma] {generated/kw51_sweep.csv};
\addplot[vit] table[x=k, y=spec, col sep=comma] {generated/kw51_sweep.csv};
\addplot[vit, dotted] table[x=k, y=sens, col sep=comma] {generated/kw51_sweep.csv};
\draw[gray] ({axis cs:2.5,0}|-{rel axis cs:0,0}) -- ({axis cs:2.5,0}|-{rel axis cs:0,1});
\legend{PCA spec., PCA sens., ViT spec., ViT sens.}
\nextgroupplot[title={V\"anersborg}]
\addplot[pca] table[x=k, y=spec_pca, col sep=comma] {generated/vanersborg_sweep.csv};
\addplot[pca, dotted] table[x=k, y=sens_pca, col sep=comma] {generated/vanersborg_sweep.csv};
\addplot[vit] table[x=k, y=spec, col sep=comma] {generated/vanersborg_sweep.csv};
\addplot[vit, dotted] table[x=k, y=sens, col sep=comma] {generated/vanersborg_sweep.csv};
\draw[gray] ({axis cs:2.5,0}|-{rel axis cs:0,0}) -- ({axis cs:2.5,0}|-{rel axis cs:0,1});
\nextgroupplot[title={Chillon}]
\addplot[pca] table[x=k, y=spec_pca, col sep=comma] {generated/chillon_sweep.csv};
\addplot[vit] table[x=k, y=spec, col sep=comma] {generated/chillon_sweep.csv};
\draw[gray] ({axis cs:2.5,0}|-{rel axis cs:0,0}) -- ({axis cs:2.5,0}|-{rel axis cs:0,1});
\end{groupplot}
\datawatermark
\end{tikzpicture}
\caption{Sensitivity (dotted) and specificity (solid) on the test windows as the threshold multiplier varies, deployed detector (black) and Transformer (blue, dashed); vertical line at the default $k=2.5$.}
\label{fig:sweep}
\end{figure}
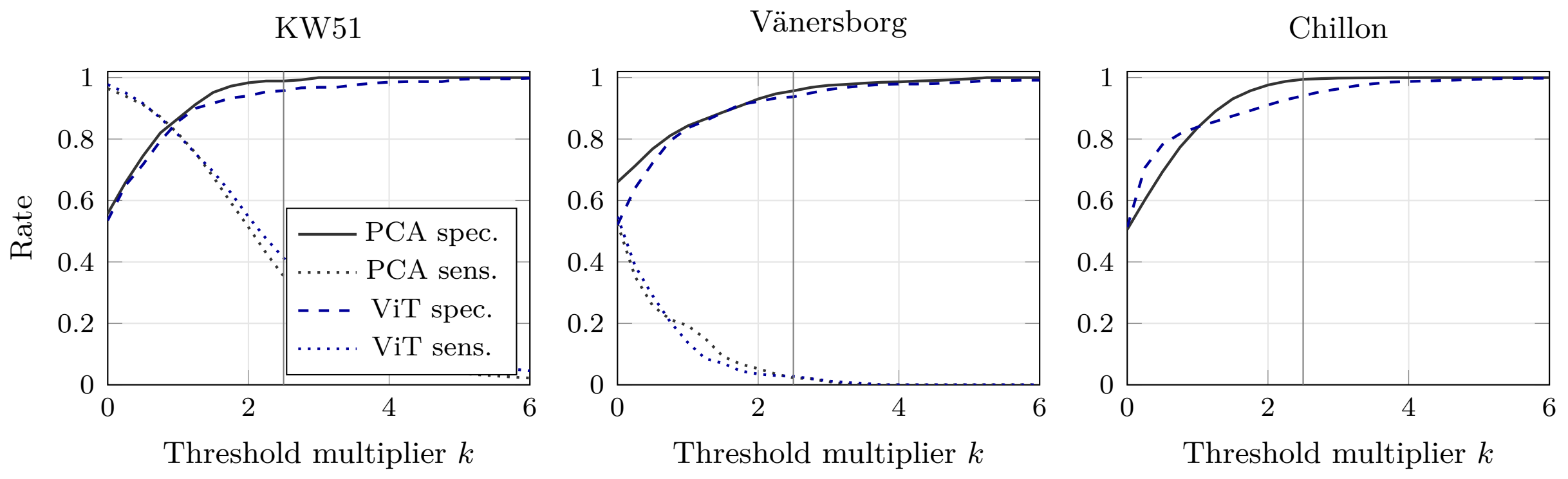

Figure~\ref{fig:sweep} reports sensitivity and specificity as the threshold multiplier $k$ in $\tau=\mathrm{median}+k\,\mathrm{sd}$ varies from 0 to 6 on the calibration windows, for both detectors. The default $k=2.5$ is marked. The curves give an owner the consequence of any other choice on these structures without refitting: on KW51 moving $k$ down buys sensitivity at a quantified specificity cost, and the deployed detector keeps more of that room than the Transformer at every $k$.

\subsection{Linear detector versus learned detector}\label{sec:cmp}
Table~\ref{tab:cmp} places the Transformer beside the deployed detector under the identical front end, training windows, calibration windows and threshold rule, in both scoring regimes. Three observations follow.

First, scored in its training regime the Transformer detects the KW51 change: AUC \aucViTKW, \sensViTKW\% sensitivity at \specViTKW\% specificity. It does not exceed the linear detector at any false-alarm rate that matters operationally: at 1\% false alarms it flags \sensAtOneViTKW\% of changed windows against \sensAtOneKW\% for PCA, at 5\% \sensAtFiveViTKW\% against \sensAtFiveKW\%, and the receiver-operating curve of the deployed detector lies on or above the Transformer's at every false-alarm rate below 30\% (Figure~\ref{fig:roc}). A model of \modelParams\ million parameters fitted on \nTrainKW\ windows is not favoured against a linear fit on a signature that is, by the documented physics, a shift of a few spectral peaks by a few bins; the result says that for this class of change the nonlinear capacity is not needed, not that it cannot be used.

Second, the scoring regime decides whether the Transformer works at all. The same weights scored on one unmasked pass give AUC \aucUnmKW\ on KW51 and \aucUnmVB\ on V\"anersborg, below chance in both cases: reference windows reconstruct slightly worse than changed windows, and the unmasked error on reference test windows (median about seven times the training loss) measures how far the input regime is from training, not how far the structure is from reference. Masked scoring removes that mismatch at the cost of four passes. The finding is general to masked autoencoders used as anomaly detectors and is the reason the released pipeline scores them as it trains them.

Third, on Chillon both detectors hold the design false-alarm rate (\specCH\% and \specViTCH\%), and on V\"anersborg neither sees the fracture (Section~\ref{sec:vb}). The deployed detector is therefore the linear one: it is at least as sensitive, it is deterministic, it fits on the node, and it costs a small fraction of the compute (Section~\ref{sec:latency}). The Transformer is retained in the released code as the learned alternative behind the same interface.

\begin{table}[t]
\centering\small
\caption{Deployed detector versus Transformer, identical front end, training windows, calibration windows and threshold rule. Rates in \%.}
\label{tab:cmp}
\footnotesize
\begin{tabular}{@{}llS[table-format=1.3]S[table-format=2.2]S[table-format=2.2]S[table-format=2.2]S[table-format=2.2]S[table-format=2.2]@{}}
\toprule
Structure & Detector & {AUC} & {Sens.} & {Spec.} & {Sens.\ at 1\%} & {Sens.\ at 5\%} & {F1} \\
\midrule
KW51 & PCA-\pcaKKW\ (deployed) & \aucKW & \sensKW & \specKW & \sensAtOneKW & \sensAtFiveKW & \fKW \\
     & MAE-ViT, masked score & \aucViTKW & \sensViTKW & \specViTKW & \sensAtOneViTKW & \sensAtFiveViTKW & \fViTKW \\
     & MAE-ViT, unmasked score & \aucUnmKW & \sensUnmKW & \specUnmKW & \sensAtOneUnmKW & \sensAtFiveUnmKW & \fUnmKW \\
\addlinespace
V\"anersborg & PCA-\pcaKVB\ (deployed) & \aucVB & \sensVB & \specVB & \sensAtOneVB & \sensAtFiveVB & \fVB \\
     & MAE-ViT, masked score & \aucViTVB & \sensViTVB & \specViTVB & \sensAtOneViTVB & \sensAtFiveViTVB & \fViTVB \\
     & MAE-ViT, unmasked score & \aucUnmVB & \sensUnmVB & \specUnmVB & \sensAtOneUnmVB & \sensAtFiveUnmVB & \fUnmVB \\
\addlinespace
Chillon & PCA-\pcaKCH\ (deployed) & {--} & {--} & \specCH & {--} & {--} & {--} \\
     & MAE-ViT, masked score & {--} & {--} & \specViTCH & {--} & {--} & {--} \\
\ifnum\vitMergedCH=1
     & MAE-ViT, unmasked score & {--} & {--} & \specUnmCH & {--} & {--} & {--} \\
\fi
\bottomrule
\end{tabular}
\end{table}

\subsection{What the V\"anersborg result bounds}\label{sec:vb}
Neither detector separates the openings before and after the fracture (Table~\ref{tab:cmp}); the AUC of \aucVB\ for the deployed detector is barely above chance and its sensitivity at the calibrated threshold (\sensVB\%) is within the false-alarm rate. This is a result about the scope of a generic spectrogram detector, and it is reported as one. The change is a crack in a single member of the counterweight truss, excited only during the opening manoeuvre, in a record that holds a few post-fracture openings; the front end was widened to 0--99.5\,Hz for this structure because the low modes carry no documented effect, and the mean-spectrum difference between the two populations is confined to a few bins between 20 and 85\,Hz with a standardized effect size near one, against sixty-second windows whose content is dominated by the manoeuvre itself. A detector that treats every channel and every minute alike, with no model of the opening sequence and no use of the sixteen strain gauges, is not the instrument for this change. What the architecture provides is the place to put a better one: the detector is a component behind a fixed interface, and an opening-aware or strain-based score for this structure class is a configuration of the same node. The published record is small enough that the result is reported with its confidence interval rather than as a definitive negative.

\subsection{Edge latency}\label{sec:latency}
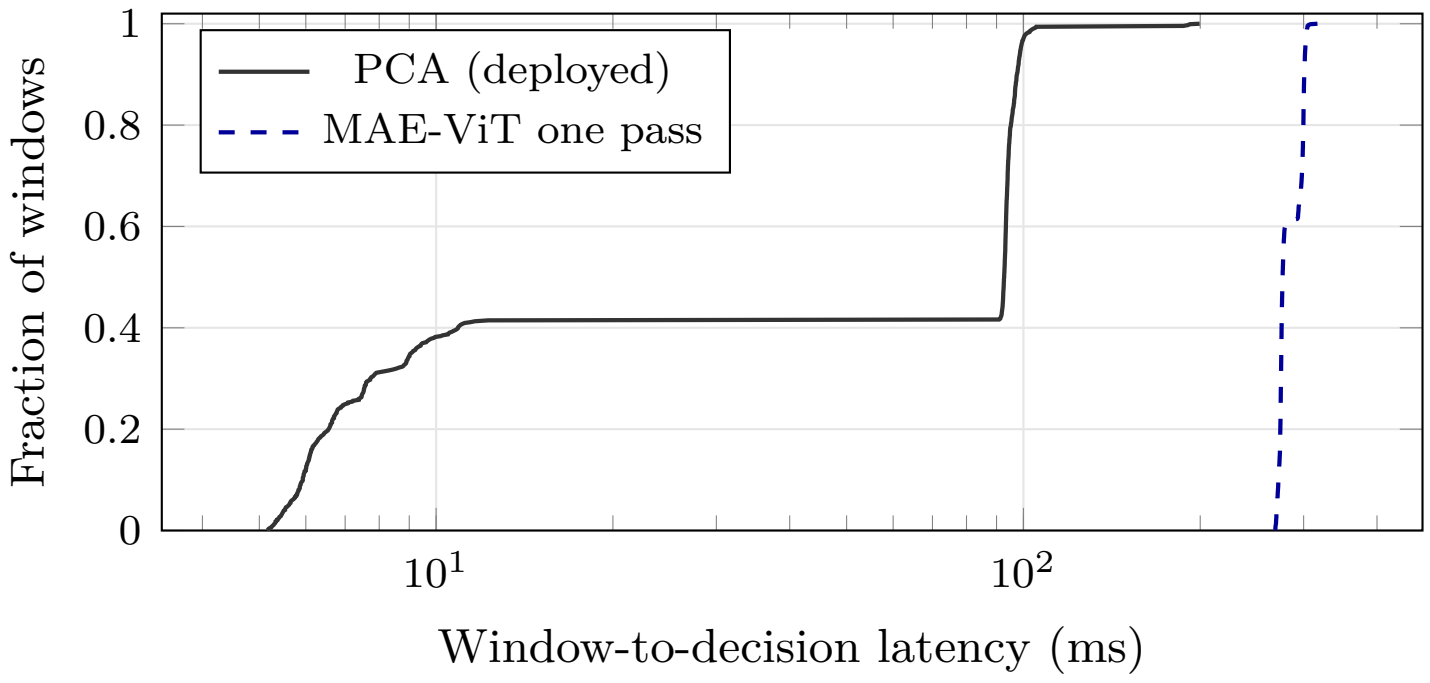
\begin{figure}[t]
\centering
\begin{tikzpicture}
\begin{semilogxaxis}[width=.62\linewidth, height=5cm, xlabel={Window-to-decision latency (ms)}, ylabel={Fraction of windows}, ymin=0, ymax=1.02, grid=major, grid style={gray!20}, legend pos=north west, legend style={font=\scriptsize}]
\addplot[pca] table[x=ms, y=cdf, col sep=comma] {generated/lat_cdf.csv};
\addplot[vit] table[x=ms, y=cdf, col sep=comma] {generated/latVit_cdf.csv};
\legend{PCA (deployed), MAE-ViT one pass}
\end{semilogxaxis}
\datawatermark
\end{tikzpicture}
\caption{Empirical distribution of window-to-decision latency in the 4-vCPU, 4\,GB container for both detectors.}
\label{fig:lat}
\end{figure}

Table~\ref{tab:lat} and Figure~\ref{fig:lat} report the window-to-decision latency measured in the 4-vCPU, 4\,GB container (\edgeHardware). \ifnum\latPcaMeasured=1 For the deployed detector the mean is \latMean\,ms, median \latMedian\,ms, P95 \latPnf\,ms, P99 \latPnn\,ms and maximum \latMax\,ms over \latn\ windows from the three structures; the spectrogram stage is \latSpec\,ms and the scoring stage \latModel\,ms, and peak resident memory is \latPeakRSSMB\,MB. \else \textcolor{red}{[PCA latency run pending: the deployed-detector figures below are a copy of the Transformer run and will be replaced by \texttt{make data}.]} \fi For the Transformer on one unmasked pass the mean is \latVitMean\,ms, median \latVitMedian\,ms, P95 \latVitPnf\,ms, P99 \latVitPnn\,ms and maximum \latVitMax\,ms, of which the forward pass is \latVitModel\,ms (\latVitModelShare\%); peak resident memory is \latVitPeakRSSMB\,MB, inside the 4\,GB budget with the operating system, and \latVitWithinTarget\% of windows complete within one second. The masked score of Section~\ref{sec:vit} costs about four times the Transformer's scoring stage. With one window per channel per minute, even the Transformer's throughput of \latVitThroughput\ windows per second serves hundreds of channels per node without a backlog. In operational terms the node reaches a decision in a fraction of a second, entirely locally; the centralized pilot cited in Section~1 needs 18--67\,s for transport alone.

\begin{table}[t]
\centering\small
\caption{Window-to-decision latency in the edge container ($n=\latn$ windows, batch size one, CPU only).}
\label{tab:lat}
\begin{tabular}{@{}lS[table-format=3.1]S[table-format=3.1]@{}}
\toprule
Latency (ms) & {PCA (deployed)} & {MAE-ViT, one pass} \\
\midrule
Mean & \latMean & \latVitMean \\
Median & \latMedian & \latVitMedian \\
P95 & \latPnf & \latVitPnf \\
P99 & \latPnn & \latVitPnn \\
Maximum & \latMax & \latVitMax \\
\midrule
Spectrogram stage (mean) & \latSpec & \latVitSpec \\
Scoring stage (mean) & \latModel & \latVitModel \\
Peak resident memory (MB) & {\latPeakRSSMB} & {\latVitPeakRSSMB} \\
\bottomrule
\end{tabular}
\end{table}

\subsection{Cross-structure summary}
Table~\ref{tab:cross} collects the properties an agency with a heterogeneous inventory needs: one pipeline and one detector family serve a steel bowstring bridge, a riveted bascule bridge and a concrete viaduct, with the structure entering only through the front-end configuration and the calibration; the per-structure calibration holds the false-alarm rate near the design level on all three; the documented global change (KW51) is detected at that false-alarm rate and the documented local change (V\"anersborg) is not.

\begin{table}[t]
\centering\small
\caption{Cross-structure summary, deployed detector.}
\label{tab:cross}
\footnotesize
\begin{tabular}{@{}p{0.3\linewidth}p{0.19\linewidth}p{0.19\linewidth}p{0.22\linewidth}@{}}
\toprule
 & KW51 & V\"anersborg & Chillon \\
\midrule
Structural form & steel bowstring & riveted steel bascule & prestressed concrete box girder \\
Documented change & connection retrofit (global) & member fracture (local) & none \\
Channels used / rate & \nChanKW\ / 100\,Hz & \nChanVB\ / 200\,Hz & \nChanCH\ / 100\,Hz \\
Reference windows train / calib.\ / test & \nTrainKW\ / \nCalibKW\ / \nNormalKW & \nTrainVB\ / \nCalibVB\ / \nNormalVB & \nTrainCH\ / \nCalibCH\ / \nNormalCH \\
AUC & \aucKW & \aucVB & -- \\
Specificity (\%) & \specKW & \specVB & \specCH \\
Sensitivity (\%) & \sensKW & \sensVB & -- \\
Front end & 0.05\,Hz, 0--10\,Hz & 0.5\,Hz, 0--100\,Hz & 0.05\,Hz, 0--10\,Hz \\
Pipeline / detector changed & no & no & no \\
\bottomrule
\end{tabular}
\end{table}

\section{Deployment Considerations}\label{sec:deploy}

\subsection{Cost projection}\label{sec:cost}
Table~\ref{tab:cost} projects annual cost for a 10{,}000-structure deployment (about 1.6\% of the 623{,}000 U.S.\ bridges) with ten sensors per structure, twenty regional hubs and published list prices as of January 2026. Edge hardware (a 4\,GB single-board computer, enclosure, power, modem and four hours of installation, amortized over five years) is 51\% of the total; cloud infrastructure is 1.2\%, a direct consequence of processing at the edge. The projection is approximately \$224 per bridge per year, with a sensitivity range of about \$150 (50{,}000 structures) to \$450 (1{,}000 structures), and \$282 if edge hardware costs 50\% more. Routine visual inspection is estimated at \$2{,}500--\$7{,}500 per bridge per year annualized across inspection types. Continuous monitoring supplements rather than replaces inspection under 23 CFR 650.311; its value is condition information between inspections that supports the risk-based intervals the 2022 standards introduce.

\begin{table}[t]
\centering\small
\caption{Annual cost projection, 10{,}000-structure deployment (USD).}
\label{tab:cost}
\begin{tabular}{@{}lrrr@{}}
\toprule
Category & Annual & Per bridge & Share \\
\midrule
Edge hardware, amortized 5 yr & 1,150,000 & 115.00 & 51.3\% \\
Cellular connectivity (enterprise IoT rate) & 31,200 & 3.12 & 1.4\% \\
Fog infrastructure, 20 hubs & 165,360 & 16.54 & 7.4\% \\
Cloud infrastructure & 26,640 & 2.66 & 1.2\% \\
Operations staff, 3 FTE & 450,000 & 45.00 & 20.1\% \\
Hardware replacement, 10\%/yr & 275,000 & 27.50 & 12.3\% \\
Software licensing & 50,000 & 5.00 & 2.2\% \\
Contingency, 10\% & 91,820 & 9.18 & 4.1\% \\
\midrule
Total & 2,240,020 & 224.00 & 100\% \\
\bottomrule
\end{tabular}
\end{table}

\subsection{Integration with inspection practice}
Edge-produced health indicators and anomaly events are designed to enter existing asset-management workflows: a flagged structure is prioritized for inspection, a quiet one supports a longer risk-based interval. The architecture is independent of sensor vendor, cloud provider and inspection software; the fog and cloud tiers are defined by their message contracts, and an agency can implement them on any platform.

\section{Scope and Validation Roadmap}\label{sec:scope}
The results establish the design on three in-service bridges and bound what its generic detector sees. The validation program continues in three phases. First, the full fifteen-month KW51 record, which places the retrofit inside a continuous series and exposes the detector to every season, together with the progressive-damage benchmarks (Z24) and nonlinear references (Isolation Forest, one-class SVM, convolutional autoencoder); for V\"anersborg, an opening-aware score and the strain-gauge channels behind the same detector interface. Second, a pilot on 10--20 instrumented structures over 6--12 months on the target single-board hardware, to measure false-alarm rates, calibration behaviour across seasons and actual operating cost. Third, expansion to 100 or more structures to exercise the fog and cloud tiers at scale. Each phase provides the measured values that replace the projected ones in Section~\ref{sec:cost}.

\section{Conclusion}\label{sec:conc}
This paper presented a reference architecture for structural health monitoring that performs anomaly detection at the structure on commodity hardware, transmits compact summaries instead of waveforms, and continues operating without network connectivity. Structural dynamics enters the edge pipeline through a per-structure spectrogram front end; the detector is a reconstruction-error model trained on reference-condition windows and calibrated per structure, behind an interface that admits a linear or a learned implementation. On openly licensed records of three in-service bridges the deployed linear detector separates the pre- and post-retrofit condition of a steel bowstring railway bridge (AUC \aucKW, \sensKW\% sensitivity at \specKW\% specificity), holds \specCH\% specificity on a prestressed concrete viaduct through a \tempMinCH--\tempMaxCH\,\si{\celsius} season, and does not see a local fracture in one member of a riveted bascule bridge (AUC \aucVB), a bound that is reported rather than removed. A \modelParams-million-parameter masked-autoencoding Vision Transformer under the identical protocol reaches AUC \aucViTKW\ on the first structure when scored in its training regime, below the linear detector, and fails when scored unmasked, a finding that applies to masked autoencoders used as anomaly detectors generally. In a 4-vCPU, 4\,GB container the node decides in \latMean\,ms on average with the deployed detector and \latVitMean\,ms with the Transformer. The projected cost of about \$224 per bridge per year at 10{,}000 structures places continuous monitoring within reach of inventory-scale deployment. The architecture, the fitted detectors, the trained weights and the scripts that regenerate every number in this paper from the public records are released so that transportation agencies, engineering firms and research groups can implement and verify the design on their own inventories.

\section*{Data and code availability}
The three datasets are published by their authors under open licenses \citep{maes2020data,leander2023data,martinsanz2019data}. The download scripts, preprocessing, front-end configurations, detectors, training configuration, per-window outputs, diagnostics and the scripts that generate every table and figure in this paper are released at \url{https://github.com/LorenYan98/edge-shm-reference-architecture} (version 1.0.0, archived at Zenodo, DOI \href{https://doi.org/10.5281/zenodo.24517852}{10.5281/zenodo.24517852}). Trained weights are regenerated by the released scripts.

\section*{Funding and competing interests}
This research received no external funding and was conducted independently of the author's employer. The author declares no competing interests.

\appendix
\section{Training and hardware details}\label{app:train}
\begin{figure}[h]
\centering
\begin{tikzpicture}
\begin{semilogyaxis}[width=.6\linewidth, height=4.6cm, xlabel={Epoch}, ylabel={Masked MSE}, grid=major, grid style={gray!20}, legend style={font=\scriptsize}]
\addplot[pca, mark=*, mark size=1.2pt] table[x=epoch, y=loss, col sep=comma] {generated/train_curve_kw51.csv};
\addplot[vit, mark=square*, mark size=1.2pt] table[x=epoch, y=loss, col sep=comma] {generated/train_curve_vanersborg.csv};
\addplot[unm, thick, mark=triangle*, mark size=1.4pt] table[x=epoch, y=loss, col sep=comma] {generated/train_curve_chillon.csv};
\legend{KW51, V\"anersborg, Chillon}
\end{semilogyaxis}
\datawatermark
\end{tikzpicture}
\caption{Training loss (mean squared error on masked patches) per epoch, one Transformer per structure.}
\label{fig:curve}
\end{figure}
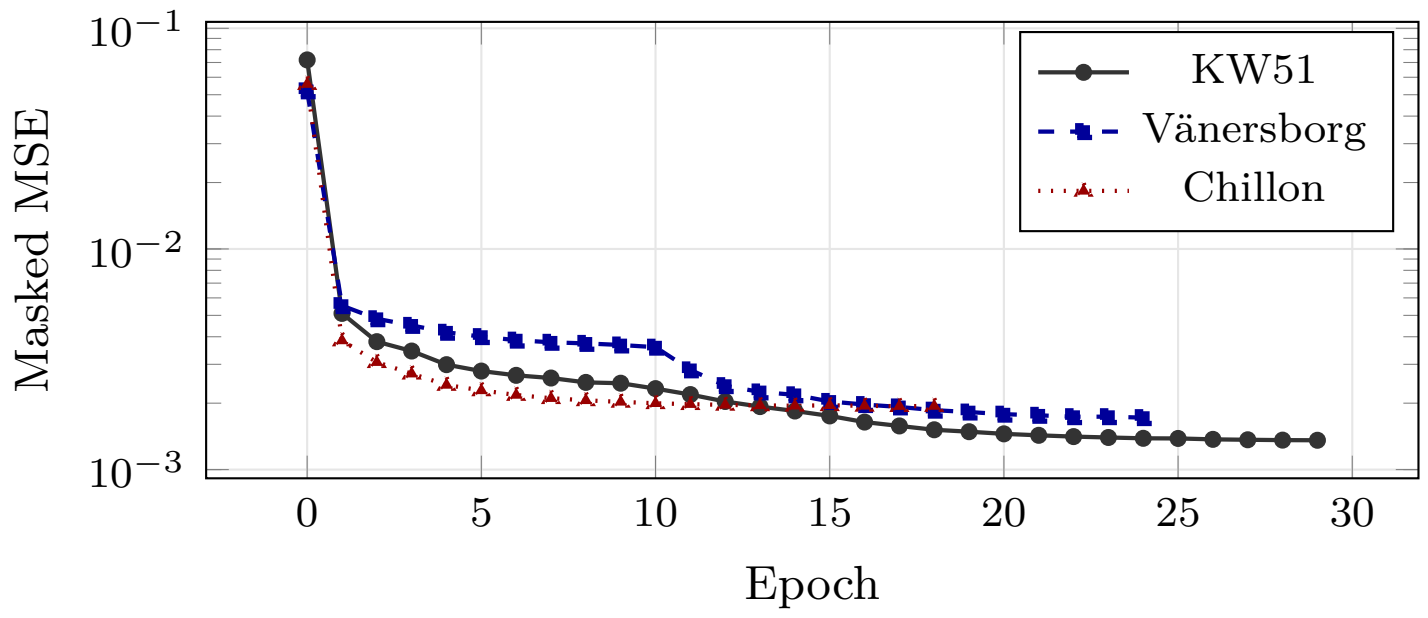

One Transformer was trained per structure on a 32-core CPU, three in parallel at ten threads each: KW51 for \epochsKW\ epochs (\trainHoursKW\,h) on \nTrainKW\ windows, V\"anersborg for \epochsVB\ epochs (\trainHoursVB\,h) on \nTrainVB, Chillon for \epochsCH\ epochs (\trainHoursCH\,h) on \nTrainCH; Figure~\ref{fig:curve} shows the masked reconstruction loss per epoch. Edge measurements used a container limited to 4 vCPUs and 4\,GB (\texttt{torch} \torchVersion, \edgeThreads\ threads). Minimum edge requirements: 4\,GB RAM, ARM or x86 CPU, 200\,MB storage for model and software, 5--10\,W, any uplink (cellular, Wi-Fi, Ethernet). The deployed PCA detector needs about 100\,MB for its components at the present image size; INT8 quantization, an optimized operator runtime and, where present, an on-board GPU are the available accelerations for the Transformer, deferred until the detection methodology is fixed.

\bibliographystyle{unsrtnat}
\bibliography{refs}

@techreport{asce2021,
  author = {{American Society of Civil Engineers}},
  title = {2021 Report Card for America's Infrastructure},
  institution = {ASCE}, year = {2021}
}

@book{farrar2012,
  author = {Farrar, Charles R. and Worden, Keith},
  title = {Structural Health Monitoring: A Machine Learning Perspective},
  publisher = {John Wiley \& Sons}, year = {2012}
}

@article{worden2007,
  author = {Worden, K. and Farrar, C. R. and Manson, G. and Park, G.},
  title = {The Fundamental Axioms of Structural Health Monitoring},
  journal = {Proceedings of the Royal Society A}, volume = {463}, number = {2082}, pages = {1639--1664}, year = {2007}
}

@techreport{doebling1996,
  author = {Doebling, S. W. and Farrar, C. R. and Prime, M. B. and Shevitz, D. W.},
  title = {Damage Identification and Health Monitoring of Structural and Mechanical Systems from Changes in Their Vibration Characteristics: A Literature Review},
  institution = {Los Alamos National Laboratory}, number = {LA-13070-MS}, year = {1996}
}

@techreport{sohn2003,
  author = {Sohn, H. and Farrar, C. R. and Hemez, F. M. and Shunk, D. D. and Stinemates, D. W. and Nadler, B. R. and Czarnecki, J. J.},
  title = {A Review of Structural Health Monitoring Literature: 1996--2001},
  institution = {Los Alamos National Laboratory}, number = {LA-13976-MS}, year = {2003}
}

@article{cawley2018,
  author = {Cawley, Peter},
  title = {Structural Health Monitoring: Closing the Gap Between Research and Industrial Deployment},
  journal = {Structural Health Monitoring}, volume = {17}, number = {5}, pages = {1225--1244}, year = {2018}
}

@book{brincker2015,
  author = {Brincker, Rune and Ventura, Carlos},
  title = {Introduction to Operational Modal Analysis},
  publisher = {John Wiley \& Sons}, year = {2015}
}

@article{cross2011,
  author = {Cross, E. J. and Worden, K. and Chen, Q.},
  title = {Cointegration: A Novel Approach for the Removal of Environmental Trends in Structural Health Monitoring Data},
  journal = {Proceedings of the Royal Society A}, volume = {467}, number = {2133}, pages = {2712--2732}, year = {2011}
}

@article{sohn2007,
  author = {Sohn, Hoon},
  title = {Effects of Environmental and Operational Variability on Structural Health Monitoring},
  journal = {Philosophical Transactions of the Royal Society A}, volume = {365}, number = {1851}, pages = {539--560}, year = {2007}
}

@article{salawu1997,
  author = {Salawu, O. S.},
  title = {Detection of Structural Damage Through Changes in Frequency: A Review},
  journal = {Engineering Structures}, volume = {19}, number = {9}, pages = {718--723}, year = {1997}
}

@article{avci2021,
  author = {Avci, O. and Abdeljaber, O. and Kiranyaz, S. and Hussein, M. and Gabbouj, M. and Inman, D. J.},
  title = {A Review of Vibration-Based Damage Detection in Civil Structures: From Traditional Methods to Machine Learning and Deep Learning Applications},
  journal = {Mechanical Systems and Signal Processing}, volume = {147}, pages = {107077}, year = {2021}
}

@article{pathirage2018,
  author = {Pathirage, C. S. N. and Li, J. and Li, L. and Hao, H. and Liu, W. and Ni, P.},
  title = {Structural Damage Identification Based on Autoencoder Neural Networks and Deep Learning},
  journal = {Engineering Structures}, volume = {172}, pages = {13--28}, year = {2018}
}

@article{bao2019,
  author = {Bao, Y. and Tang, Z. and Li, H. and Zhang, Y.},
  title = {Computer Vision and Deep Learning-Based Data Anomaly Detection Method for Structural Health Monitoring},
  journal = {Structural Health Monitoring}, volume = {18}, number = {2}, pages = {401--421}, year = {2019}
}

@article{abdeljaber2017,
  author = {Abdeljaber, O. and Avci, O. and Kiranyaz, S. and Gabbouj, M. and Inman, D. J.},
  title = {Real-Time Vibration-Based Structural Damage Detection Using One-Dimensional Convolutional Neural Networks},
  journal = {Journal of Sound and Vibration}, volume = {388}, pages = {154--170}, year = {2017}
}

@article{cha2017,
  author = {Cha, Y. J. and Choi, W. and B{\"u}y{\"u}k{\"o}zt{\"u}rk, O.},
  title = {Deep Learning-Based Crack Damage Detection Using Convolutional Neural Networks},
  journal = {Computer-Aided Civil and Infrastructure Engineering}, volume = {32}, number = {5}, pages = {361--378}, year = {2017}
}

@inproceedings{he2022,
  author = {He, K. and Chen, X. and Xie, S. and Li, Y. and Doll{\'a}r, P. and Girshick, R.},
  title = {Masked Autoencoders Are Scalable Vision Learners},
  booktitle = {IEEE/CVF Conference on Computer Vision and Pattern Recognition (CVPR)}, pages = {16000--16009}, year = {2022}
}

@inproceedings{dosovitskiy2021,
  author = {Dosovitskiy, A. and Beyer, L. and Kolesnikov, A. and Weissenborn, D. and Zhai, X. and Unterthiner, T. and Dehghani, M. and Minderer, M. and Heigold, G. and Gelly, S. and Uszkoreit, J. and Houlsby, N.},
  title = {An Image is Worth 16x16 Words: Transformers for Image Recognition at Scale},
  booktitle = {International Conference on Learning Representations (ICLR)}, year = {2021}
}

@inproceedings{vaswani2017,
  author = {Vaswani, A. and Shazeer, N. and Parmar, N. and Uszkoreit, J. and Jones, L. and Gomez, A. N. and Kaiser, {\L}. and Polosukhin, I.},
  title = {Attention Is All You Need},
  booktitle = {Advances in Neural Information Processing Systems}, volume = {30}, pages = {5998--6008}, year = {2017}
}

@article{shi2016,
  author = {Shi, W. and Cao, J. and Zhang, Q. and Li, Y. and Xu, L.},
  title = {Edge Computing: Vision and Challenges},
  journal = {IEEE Internet of Things Journal}, volume = {3}, number = {5}, pages = {637--646}, year = {2016}
}

@inproceedings{bonomi2012,
  author = {Bonomi, F. and Milito, R. and Zhu, J. and Addepalli, S.},
  title = {Fog Computing and Its Role in the Internet of Things},
  booktitle = {Proceedings of the First Edition of the MCC Workshop on Mobile Cloud Computing}, pages = {13--16}, year = {2012}
}

@article{hackmann2014,
  author = {Hackmann, G. and Guo, W. and Yan, G. and Sun, Z. and Lu, C. and Dyke, S.},
  title = {Cyber-Physical Codesign of Distributed Structural Health Monitoring with Wireless Sensor Networks},
  journal = {IEEE Transactions on Parallel and Distributed Systems}, volume = {25}, number = {1}, pages = {63--72}, year = {2014}
}

@article{lynch2006,
  author = {Lynch, J. P. and Loh, K. J.},
  title = {A Summary Review of Wireless Sensors and Sensor Networks for Structural Health Monitoring},
  journal = {Shock and Vibration Digest}, volume = {38}, number = {2}, pages = {91--130}, year = {2006}
}

@article{sony2019,
  author = {Sony, S. and Laventure, S. and Sadhu, A.},
  title = {A Literature Review of Next-Generation Smart Sensing Technology in Structural Health Monitoring},
  journal = {Structural Control and Health Monitoring}, volume = {26}, number = {3}, pages = {e2321}, year = {2019}
}

@article{zhang2024transformer,
  author = {Zhang, Qingyu and Guo, Maozu and Zhao, Lingling and Li, Yang and Zhang, Xinxin and Han, Miao},
  title = {Transformer-Based Structural Seismic Response Prediction},
  journal = {Structures}, volume = {61}, pages = {105929}, year = {2024},
  doi = {10.1016/j.istruc.2024.105929}
}

@article{li2019fcn,
  author = {Li, S. and Zhao, X. and Zhou, G.},
  title = {Automatic Pixel-Level Multiple Damage Detection of Concrete Structure Using Fully Convolutional Network},
  journal = {Computer-Aided Civil and Infrastructure Engineering}, volume = {34}, number = {7}, pages = {616--634}, year = {2019}
}

@book{kleppmann2017,
  author = {Kleppmann, Martin},
  title = {Designing Data-Intensive Applications},
  publisher = {O'Reilly Media}, year = {2017}
}

@book{nygard2018,
  author = {Nygard, Michael T.},
  title = {Release It! Design and Deploy Production-Ready Software},
  edition = {2nd}, publisher = {Pragmatic Bookshelf}, year = {2018}
}

@article{buckley2021,
  author = {Buckley, Tadhg and Ghosh, Bidisha and Pakrashi, Vikram},
  title = {Edge Structural Health Monitoring ({E-SHM}) Using Low-Power Wireless Sensing},
  journal = {Sensors}, volume = {21}, number = {20}, pages = {6760}, year = {2021},
  doi = {10.3390/s21206760}
}

@article{hidalgofort2024,
  author = {Hidalgo-Fort, Eduardo and Blanco-Carmona, Pedro and Mu{\~n}oz-Chavero, Fernando and Torralba, Antonio and Castro-Triguero, Rafael},
  title = {Low-Cost, Low-Power Edge Computing System for Structural Health Monitoring in an {IoT} Framework},
  journal = {Sensors}, volume = {24}, number = {15}, pages = {5078}, year = {2024},
  doi = {10.3390/s24155078}
}

@article{ribeiro2023,
  author = {Ribeiro, Renan Rocha and Sobral, Rafael de Almeida and Cavalcante, Ian Barreto and Veloso, Lu{\'i}s Augusto Conte Mendes and Lameiras, Rodrigo de Melo},
  title = {A Low-Cost Wireless Multinode Vibration Monitoring System for Civil Structures},
  journal = {Structural Control and Health Monitoring}, volume = {2023}, pages = {5240059}, year = {2023},
  doi = {10.1155/2023/5240059}
}

@misc{zhang2025edge,
  author = {Zhang, Yuxuan and Xu, Ye and Martinez-Rau, Luciano Sebastian and Vu, Quynh Nguyen Phuong and Oelmann, Bengt and Bader, Sebastian},
  title = {On-Device Crack Segmentation for Edge Structural Health Monitoring},
  year = {2025}, eprint = {2505.07915}, archivePrefix = {arXiv}, primaryClass = {cs.LG},
  howpublished = {arXiv:2505.07915}
}

@misc{mishra2023,
  author = {Mishra, Anoop and Gangisetti, Gopinath and Khazanchi, Deepak},
  title = {Integrating {Edge-AI} in Structural Health Monitoring Domain},
  year = {2023}, eprint = {2304.03718}, archivePrefix = {arXiv}, primaryClass = {cs.LG},
  howpublished = {arXiv:2304.03718}
}

@article{martin2022,
  author = {Mart{\'i}n, Cristian and Garrido, Daniel and Llopis, Luis and Rubio, Bartolom{\'e} and D{\'i}az, Manuel},
  title = {Facilitating the Monitoring and Management of Structural Health in Civil Infrastructures with an {Edge/Fog/Cloud} Architecture},
  journal = {Computer Standards \& Interfaces}, volume = {81}, pages = {103600}, year = {2022},
  doi = {10.1016/j.csi.2021.103600}
}

@misc{nbis2022,
  author = {{Federal Highway Administration}},
  title = {National Bridge Inspection Standards; Final Rule, 23 CFR Part 650 Subpart C},
  howpublished = {Federal Register 87 FR 27396}, year = {2022}
}

@misc{benfenati2024,
  author = {Benfenati, Luca and Jahier Pagliari, Daniele and Zanatta, Luca and Bedoya Velez, Yhorman Alexander and Acquaviva, Andrea and Poncino, Massimo and Macii, Enrico and Benini, Luca and Burrello, Alessio},
  title = {Foundation Models for Structural Health Monitoring},
  year = {2024}, eprint = {2404.02944}, archivePrefix = {arXiv}, primaryClass = {cs.LG},
  howpublished = {arXiv:2404.02944}
}

@article{maes2021,
  author = {Maes, Kristof and Lombaert, Geert},
  title = {Monitoring Railway Bridge {KW51} Before, During, and After Retrofitting},
  journal = {Journal of Bridge Engineering}, volume = {26}, number = {3}, pages = {04721001}, year = {2021},
  doi = {10.1061/(ASCE)BE.1943-5592.0001668}
}

@misc{maes2020data,
  author = {Maes, Kristof and Lombaert, Geert},
  title = {Monitoring data for railway bridge {KW51} in {Leuven}, {Belgium}, before, during, and after retrofitting},
  year = {2020}, publisher = {Zenodo}, doi = {10.5281/zenodo.3745914}, note = {CC BY-NC-SA 4.0}
}

@article{leander2023,
  author = {Leander, John and Nyman, Jacob and Karoumi, Raid and Rosengren, Peter and Johansson, Gunnar},
  title = {Dataset for damage detection retrieved from a monitored bridge pre and post verified damage},
  journal = {Data in Brief}, volume = {51}, pages = {109729}, year = {2023}, doi = {10.1016/j.dib.2023.109729}
}

@misc{leander2023data,
  author = {Leander, John and Nyman, Jacob and Karoumi, Raid and Rosengren, Peter and Johansson, Gunnar},
  title = {Dataset from structural health monitoring of a steel bridge in {Sweden}},
  year = {2023}, publisher = {Zenodo}, doi = {10.5281/zenodo.8300495}, note = {CC BY 4.0}
}

@article{martinsanz2020,
  author = {Mart{\'i}n-Sanz, Henar and Tatsis, Konstantinos and Dertimanis, Vasilis K. and Avenda{\~n}o-Valencia, Luis David and Br{\"u}hwiler, Eugen and Chatzi, Eleni},
  title = {Monitoring of the {UHPFRC} strengthened {Chillon} viaduct under environmental and operational variability},
  journal = {Structure and Infrastructure Engineering}, volume = {16}, number = {1}, pages = {138--168}, year = {2020},
  doi = {10.1080/15732479.2019.1650079}
}

@misc{martinsanz2019data,
  author = {Mart{\'i}n-Sanz, Henar and Tatsis, Konstantinos and Dertimanis, Vasilis K. and Avenda{\~n}o-Valencia, Luis David and Br{\"u}hwiler, Eugen and Chatzi, Eleni},
  title = {Monitoring of the {UHPFRC} strengthened {Chillon} viaduct under environmental and operational variability (dataset)},
  year = {2019}, publisher = {Zenodo}, doi = {10.5281/zenodo.3234805}, note = {CC BY 4.0}
}
\end{document}